\documentclass{tlp}
\usepackage{graphicx}
\usepackage{proof}
\usepackage{latexsym}
\usepackage{amssymb}
\usepackage{amsfonts}
\usepackage{epic,eepic}
\usepackage{mysym}

\newcommand{\Exp}{\calE}
\newcommand{\CExp}{{\cal E}_{\mathrm{ch}}}

\newcommand {\MSRC} {MSR(${\cal C}$)}

\newcommand{\Sol}{Sol}

\newcommand{\arrowup}[3]{\setbox0=\hbox{$\ {}^{#2}\ $}
  \setbox1=\hbox{$\longrightarrow$}
  \ifdim\wd0<\wd1\setbox0=\box1\else\relax\fi
  {#1}\,\mathop{\hbox to \wd0{\rightarrowfill}}\limits^{#2}\,{#3}
}

\newcommand{\calA}{ {\cal A} }
\newcommand{\calB}{ {\cal B} }
\newcommand{\calC}{ {\cal C} }
\newcommand{\calE}{ {\cal E} }
\newcommand{\calI}{ {\cal I} }
\newcommand{\calH}{ {\cal H} }

\newcommand{\calP}{ {\cal P} }

\newcommand{\calR}{ {\cal R} }
\newcommand{\calS}{ {\cal S} }
\newcommand{\calT}{ {\cal T} }
\newcommand{\calL}{ {\cal L} }
\newcommand{\calM}{ {\cal M} }
\newcommand{\calN}{ {\cal N} }
\newcommand{\calV}{ {\cal V} }

\newcommand{\Rat}{ \mathbb{Q} }

\newcommand{\bfS} { {S} }
\newcommand{\bfPre} { {SPre} }

\newcommand{\den}[1] { [\![{#1}]\!]}
\newcommand{\comment}[1]{}
\newcommand{\tuple}[1]{ \langle {#1} \rangle}
\newcommand{\lguard}{[}
\newcommand{\rguard}{]}
\newcommand{\TDLtoMSR}[1]{#1^\bullet}
\newtheorem {definition}{Definition}
\newtheorem {proposition}{Proposition}

\newtheorem {theorem}{Theorem}
\newtheorem {corollary}{Corollary}
\newtheorem {es}{Example}
\newenvironment{example}{\begin{es}\rm}{\hspace*{\fill}$\Box$\end{es}}

\title[Constraint-based Verification of Abstract Multithreaded Programs]
{Constraint-based Automatic Verification of Abstract Models of Multithreaded Programs}
\author[Giorgio Delzanno]
{GIORGIO DELZANNO\\
\begin{tabular}{c}
Dipartimento di Informatica e Scienze dell'Informazione,
Universit\`a di Genova\\
via Dodecaneso 35, 16146 Genova - Italy\\
\email{giorgio@disi.unige.it}
\end{tabular}
}

\pagerange{\pageref{firstpage}--\pageref{lastpage}}

\volume{}
\jdate{}
\pubyear{}

\submitted{17 December 2003}
\revised{13 April 2005}
\accepted{15 January 2006}

\begin{document}

\label{firstpage}

\maketitle

\begin{abstract}
We present a technique for the automated verification of
 {abstract models of multithreaded programs} providing {fresh name generation},
{name mobility}, and {unbounded control}.

As high level specification language we adopt here an extension of
communication finite-state machines with local variables ranging over
an infinite name domain, called TDL programs.
Communication machines have been proved very effective for representing
communication protocols as well as for representing abstractions of
multithreaded software.

The verification method that we propose is based on the encoding of  TDL programs
into a low level language based on  {multiset rewriting} and
{constraints} that can be viewed as  an extension of Petri Nets.
By means of this encoding, the symbolic verification procedure developed for the
low level language in our previous work can now be applied to TDL programs.
Furthermore, the encoding  allows us to isolate a decidable class of
verification problems for TDL programs that still provide
{fresh name generation}, {name mobility}, and {unbounded control}.
Our syntactic restrictions are in fact defined on the internal structure of
threads: In order to obtain a complete and terminating method,
threads are only allowed to have at most one local variable (ranging over an
infinite domain of names).
\end{abstract}
\section{Introduction}
\label{sec:introduction}
Andrew Gordon \cite{Gor00} defines a {\em nominal calculus} to be a computational
formalism that includes a set of {\em pure names} and allows the
dynamic generation of {\em fresh}, {\em unguessable} names.
A name is {\em pure} whenever it is only useful for comparing for
identity with other names.
The use of pure names is ubiquitous in programming
languages. Some important examples are memory pointers in
imperative languages, identifiers in concurrent programming
languages, and nonces in security protocols.
In addition to pure names, a {\em nominal process calculus} should
provide mechanisms for {\em concurrency} and {\em inter-process
communication}. A computational model that  provides all these features
is an adequate abstract formalism for the analysis of
{\em multithreaded} and {\em distributed} software.
\paragraph{The Problem}
Automated verification of specifications in a
nominal process calculus becomes particularly challenging in presence of
the following three features:
the possibility of generating fresh names ({\em name generation});
the possibility of transmitting names ({\em name mobility});
the possibility of dynamically adding new threads of control
({\em unbounded control}).
In fact, a calculus that provides all the previous features can be used
to specify systems with a state-space infinite in {\em several dimensions}.
This feature makes difficult (if not impossible) the application of
finite-state verification techniques or techniques based on abstractions
of process specifications into Petri Nets or CCS-like models.
In recent years there have been several attempts of extending automated verification
methods from finite-state to infinite-state systems \cite{AN00,Pnueli}.
In this paper we are interested in investigating the possible application
of the methods we proposed in \cite{Del01} to verification problems of interest for
nominal process calculi.
\paragraph{Constraint-based Symbolic Model Checking}
In \cite{Del01} we introduced a specification language,
called MSR($\calC$), for the analysis of  communication protocols whose
specifications are parametric in several dimensions (e.g. number of servers, clients,
and tickets as in the model of the ticket mutual exclusion algorithm shown in \cite{BD02}).
MSR($\calC$) combines multiset rewriting over first order atomic formulas
\cite{CDLMS99} with constraints programming. More specifically,
multiset rewriting is used to specify the {control part} of a concurrent
system, whereas {constraints} are used to symbolically specify the relations
over local data.
The verification method  proposed in \cite{Del02} allows
us to symbolically reason on the behavior of MSR($\calC$) specifications.
To this aim, following \cite{ACJT96,AN00} we introduced  a symbolic representation
of infinite collections of {global configurations} based on the combination
of multisets of atomic formulas and constraints, called constrained
configurations.\footnote{Notice that in  \cite{ACJT96,AN00} a {\em constraint}
denotes a symbolic state whereas we use the word {\em constraint} to denote a symbolic
representation of the relation of data variables (e.g. a linear arithmetic formula)
used as part of the symbolic representation of sets of states (a constrained
configuration).}
The verification procedure performs a  symbolic backward reachability analysis
by means of a symbolic {\em pre-image}  operator that works over constrained
configurations  \cite{Del02}.
The main feature of this method is the possibility of automatically
handling systems with an arbitrary number of components.
Furthermore, since we use a symbolic and finite representation of possibly
infinite sets of configurations, the analysis is carried out without loss
of precision.
\smallskip\\
A natural question for our research is whether and how these techniques
can be used for verification of abstract models of multithreaded programs.
\paragraph{Our Contribution}
In this paper we propose a sound, and fully automatic verification method
for abstract models of multithreaded programs that provide {\em name generation},
{\em name mobility},  and {\em unbounded control}.
As a high level specification language we adopt here an extension with value-passing
of the formalism of \cite{BCR01} based on families of state machines used to specify
abstractions of multithreaded software libraries.
The resulting language is called Thread Definition Language (TDL).
This formalism allows us to keep separate the {finite control component} of a
{thread definition} from the management of {local variables} (that in our
setting range over a infinite set of names), and to treat in isolation
the operations to generate fresh names, to transmit names, and
to create new threads.
In the present paper we will show that the extension of the model of
\cite{BCR01} with value-passing makes the model Turing equivalent.

The verification methodology is based on the encoding of TDL programs
into a specification in the instance MSR$_{NC}$  of the language scheme
\MSRC~of\cite{Del01}.
MSR$_{NC}$ is obtained by taking as constraint system a subclass of linear
arithmetics with only $=$ and $>$ relations between variables, called name
constraints ($NC$).
The low level specification language MSR$_{NC}$ is not just instrumental for
the encoding of  TDL programs. Indeed, it has
been applied to model consistency and mutual exclusion protocols
in \cite{BD02,Del02}.
Via this encoding, the verification method based on
symbolic backward reachability obtained by instantiating the general method
for \MSRC~to NC-constraints can now be applied to abstract models of multithreaded
programs.
Although termination is not guaranteed in general, the resulting verification method
can succeed on practical  examples as the Challenge-Response TDL program
defined over binary predicates we will illustrated in the present paper.
Furthermore, by propagating the sufficient conditions for
termination defined in \cite{BD02,Del02} back to TDL programs,
we obtain an interesting class of decidable problems for
abstract models of multithreaded programs still providing name generation,
name mobility, and unbounded control.
\paragraph{Plan of the Paper}%
In Section \ref{NominalCalculus} we present the Thread Definition Language (TDL)
with examples of multithreaded programs. Furthermore, we discuss the expressiveness
of TDL programs showing that they can simulate Two Counter Machines.
In Section \ref{MSRTranslation}, after introducing the MSR$_{NC}$ formalism,
we show that  TDL programs can be simulated by  MSR$_{NC}$ specifications.
In Section \ref{Verification} we show how to transfer the verification methods
developed for \MSRC~to TDL programs. Furthermore, we show that safety properties
can be decided for the special class of monadic TDL programs.
In Section \ref{conclusions} we address some conclusions and discuss related work.
\section{Thread Definition Language (TDL)}
\label{NominalCalculus}
In this section we will define TDL programs.
This formalism is a natural extension with value-passing
of the communicating machines used by \cite{BCR01} to specify
abstractions of multithreaded software libraries.
\paragraph{Terminology}
Let $\calN$ be a denumerable set of {\em names} equipped with
the relations $=$ and $\neq$ and a special element $\bot$
such that $n\neq \bot$ for any $n\in\calN$.
Furthermore, let $\calV$ be a denumerable set of variables,
$\calC=\{c_1,\ldots,c_m\}$ a finite set of
constants, and $\calL$ a finite set of {\em internal action} labels.
For a fixed $V\subseteq\calV$, the set of {\em expressions} is defined as
$\Exp=V\cup\calC\cup\{\bot\}$ (when necessary we will use $\Exp(V)$ to explicit the set
of variables $V$ upon which expressions are defined).
The set of {\em channel expressions}  is defined as $\CExp=V\cup \calC$.
Channel expressions will be used as synchronization labels so as to establish
communication links only at execution time.
\smallskip\\
A {\em guard over $V$} is a conjunction  $\gamma_1,\ldots,\gamma_s$,
where $\gamma_i$ is either $true$, $x=e$ or $x\neq e$ with $x\in V$ and $e\in\Exp$
for $i:1,\ldots,s$.
An {\em assignment} $\alpha$ from $V$ to $W$ is a conjunction
like $x_i:=e_i$ where $x_i\in W$,
$e_i\in\Exp(V)$ for $i:1,\ldots k$ and $x_r\neq x_s$ for $r\neq s$.
A {\em message template $m$ over $V$} is a tuple $m=\tuple{x_1,\ldots,x_u}$
of variables in $V$.
\begin{definition}\rm
A {\em TDL program} is a set $\calT=\{P_1,\ldots,P_t\}$ of {\em thread definitions}
(with distinct names for local variables control locations).
A {\em thread definition} $P$ is a tuple $\tuple{Q,s_0,V,R}$,
where $Q$ is a finite set of {\em control locations}, $s_0\in Q$ is the initial location,
 $V\subseteq\calV$ is a finite set of {\em local variables}, and
 $R$ is a set of rules. Given $s,s'\in Q$, and $a\in\calL$,
 a {\em rule} has one of the following forms\footnote{In this paper we keep assignments, name
generation, and thread creation separate in order to simplify the
presentation of the encoding into MSR.}:
\begin{itemize}
\item[$\bullet$] {\em Internal move}: $\arrowup{s}{a}{s'}\lguard\gamma,\alpha\rguard$,
where $\gamma$ is a {\em guard over $V$}, and $\alpha$ is an
 {\em assignment} from $V$ to $V$;
\item[$\bullet$] {\em Name generation}: $\arrowup{s}{a}{s'}\lguard{x:=new}\rguard$, where
 $x\in V$, and the expression $new$ denotes a {\em fresh name};
\item[$\bullet$] {\em Thread creation:} $\arrowup{s}{a}{s'}\lguard{run~P'~with~\alpha}\rguard$,
where  $P'=\tuple{Q',t,W,R'}\in\calT$, and $\alpha$ is an {\em
assignment from $V$ to $W$} that specifies the initialization of
the local variables of the new thread;
\item[$\bullet$] {\em Message sending:} $\arrowup{s}{e!m}{s'}\lguard \gamma,\alpha\rguard$,
where $e$ is a {\em channel expression}, $m$ is a {\em message
template over $V$} that specify which names to pass,
$\gamma$ is a {\em guard over $V$}, and $\alpha$ is an {\em assignment} from $V$ to $V$.
\item[$\bullet$] {\em Message reception:} $\arrowup{s}{e?m}{s'}\lguard \gamma,\alpha\rguard$,
where  $e$ is a channel expression, $m$ is a
message template over a {\em new} set of variables $V'$ ($V'\cap V=\emptyset$)
that specifies the names to receive,
$\gamma$ is a {\em  guard over $V\cup V'$} and $\alpha$ is  an {\em  assignment}
from $V\cup V'$ to $V$.
\end{itemize}
\end{definition}
Before giving an example, we will formally introduce the operational
semantics of TDL programs.
\subsection{Operational Semantics}
In the following we will use $N$ to indicate the subset of {\em used names} of $\calN$.
Every constant $c\in\calC$ is mapped to a distinct name $n_c\neq\bot\in N$, and $\bot$
is mapped to $\bot$.
\\
Let $P=\tuple{Q,s,V,R}$ and $V=\{x_1,\ldots,x_k\}$.
A {\em local configuration} is a tuple $p=\tuple{s',n_1,\ldots,n_k}$ where $s'\in Q$ and $n_i\in N$ is the
current value of the variable $x_i\in V$ for $i:1,\ldots,k$.
\\
A {\em global configuration} $G=\tuple{N,p_{1},\ldots,p_{m}}$ is
such that $N\subseteq\calN$ and $p_{1},\ldots,p_{m}$ are local
configurations defined over $N$ and over the thread definitions in
$\calT$. Note that there is no relation between indexes in a
global configuration in $G$ and in $\calT$; $G$ is a {\em pool} of
active threads, and {\em several active threads} can be instances
of the same {\em thread definition}.
\\
Given a local configuration $p=\tuple{s',n_1,\ldots,n_k}$,
we define the {\em valuation} $\rho_p$ as
 $\rho_p(x_i)=n_i$ if $x_i\in V$,
 $\rho_p(c)=n_{c}$ if  $c\in\calC$, and
 $\rho_p(\bot)=\bot$.
 Furthermore, we say that $\rho_p$ satisfies the guard $\gamma$ if $\rho_p(\gamma)\equiv true$,
 where $\rho_p$ is extended to constraints in the natural way
 ($\rho_p(\varphi_1\wedge \varphi_2)=\rho_p(\varphi_1)\wedge\rho_p(\varphi_2)$, etc.).
\smallskip\\
The execution of $x:=e$ has the  effect of updating the local
variable $x$ of a thread with the current value of $e$ (a name
taken from the set of used values $N$). On the contrary, the
execution of $x:=new$ associates a {\em fresh unused name} to $x$.
The formula $run~P~with~\alpha$ has the effect of adding a new
thread (in its initial control location) to the current global
configuration. The initial values of the local variables of the
generated thread are determined by the execution of $\alpha$ whose
source variables are the local variables of the parent thread. The
channel names used in a rendez-vous are determined by evaluating
the channel expressions tagging sender and receiver rules. Value
passing is achieved by extending the evaluation associated to the
current configuration of the receiver so as to associate the
output message of the sender  to the variables in the input
message template. The operational semantics is given via a binary
relation $\Rightarrow$ defined  as follows.
\begin{definition}\rm
\label{sos}
Let  $G=\tuple{N,\ldots,\mathbf{p},\ldots}$, and
$\mathbf{p}=\tuple{s,n_1,\ldots,n_k}$ be a local configuration for
$P=\tuple{Q,s,V,R}$, $V=\{x_1,\ldots,x_k\}$, then:
\begin{itemize}
\item[$\bullet$] If there exists a rule $\arrowup{s}{a}{s'}\lguard\gamma,\alpha\rguard$ in $R$
such that $\rho_{\mathbf{p}}$ satisfies $\gamma$, then
 $G\Rightarrow\tuple{N,\ldots,\mathbf{p'},\ldots}$ (meaning that only $\mathbf{p}$ changes)
 where $\mathbf{p'}=\tuple{s',n_1',\ldots,n_k'}$,
   $n_i'=\rho_{\mathbf{p}}(e_i)$ if $x_i:=e_i$ is in $\alpha$,
    $n_i'=n_i$ otherwise, for $i:1,\ldots,k$.
\item[$\bullet$]
 If there exists a rule $\arrowup{s}{a}{s'}\lguard x_i:=new\rguard$ in $R$,
then
 $G\Rightarrow\tuple{N',\ldots,\mathbf{p'},\ldots}$
where
 $\mathbf{p'}=\tuple{s',n_1',\ldots,n_k'}$,
 $n_i$ is an unused name, i.e., $n_i'\in \calN\setminus N$,
 $n_j'=n_j$ for every $j\neq i$, and $N'=N\cup\{n_i'\}$;
\item[$\bullet$]
 If there exists a rule
$\arrowup{s}{a}{s'}\lguard run~P'~with~\alpha\rguard$ in $R$ with
$P'=\tuple{Q',t_0,W,R'}$, $W=\{y_{1},\ldots,y_{u}\}$, and
$\alpha$ is defined as $y_{1}:=e_{1},\ldots,y_{u}:=e_{u}$
 then
 $G\Rightarrow\tuple{N,\ldots,\mathbf{p'},\ldots,\mathbf{q}}$
 (we add a new thread whose initial local configuration is $\mathbf{q}$)
where
 $\mathbf{p'}=\tuple{s',n_1,\ldots,n_k}$, and
 $\mathbf{q}=\tuple{t_0,\rho_{\mathbf{p}}(e_{1}),\ldots,\rho_{\mathbf{p}}(e_{u})}$.
\item[$\bullet$]
 Let $\mathbf{q}=\tuple{t,m_1,\ldots,m_r}$ (distinct from $\mathbf{p}$)
 be a local configuration in $G$ associated with
 $P'=\tuple{Q',t_0,W,R'}$.\\
Let $\arrowup{s}{e!m}{s'}\lguard\gamma,\alpha\rguard$ in $R$ and
$\arrowup{t}{e'?m'}{t'}\lguard\gamma',\alpha'\rguard$ in $R'$ be
two rules such that $m=\tuple{x_1,\ldots,x_u}$,
$m'=\tuple{y_1,\ldots,y_v}$ and $u=v$ (message templates match).
We define $\sigma$ as the {\em value passing} evaluation
$\sigma(y_i)=\rho_{\mathbf{p}}(x_{i})$ for $i:1,\ldots,u$, and
$\sigma(z)=\rho_{\mathbf{q}}(z)$ for $z\in W'$.
\\
Now if $\rho_{\mathbf{p}}(e)=\rho_{\mathbf{p}}(e')$ (channel names
match), $\rho_{\mathbf{p}}$ satisfies $\gamma$, and that $\sigma$
satisfies $\gamma'$, then
 $
\tuple{N,\ldots,\mathbf{p},\ldots,\mathbf{q},\ldots}\Rightarrow
\tuple{N,\ldots,\mathbf{p'},\ldots,\mathbf{q'},\ldots}$ where
$\mathbf{p'}=\tuple{s',n_1',\ldots,n_k'}$,
$n_i'=\rho_{\mathbf{p}}(v)$ if $x_i:=v$ is in $\alpha$, $n_i'=n_i$
otherwise for $i:1,\ldots,k$;
$\mathbf{q'}=\tuple{t',m_1',\ldots,m_r'}$, $m_i'=\sigma(v)$ if
$u_i:=v$ is in $\alpha'$,
 $m_i'=m_i$ otherwise for $i:1,\ldots,r$.
\end{itemize}
\end{definition}
\begin{definition}\rm
An {\em initial global configuration} $G_0$ has an {\em arbitrary
(but finite) number} of threads with local variables all set to
$\bot$. A {\em run} is a sequence $G_0G_1\ldots$ such that
$G_{i}\Rightarrow G_{i+1}$ for $i\geq 0$. A global configuration
$G$ is {\em reachable} from $G_0$ if there exists a run from $G_0$
to $G$.
\end{definition}

\begin{example}
\label{AliceBob} Let us consider a {\em challenge and
response} protocol in which the goal of two agents Alice and Bob
is to exchange a pair of new names $\tuple{n_A,n_B}$,
the first one created by Alice and the second one created by Bob,
so as to build a composed secret key.
We can specify the protocol by using new names
to dynamically establish {\em private channel names} between
instances of the initiator and of the responder.
The TDL program in Figure \ref{model} follows this idea.
The thread $Init$ specifies the behavior of the initiator.
He first creates a new name using the internal action $fresh$, and stores it
in the local variable  $n_A$.
Then, he sends $n_A$ on channel $c$ (a constant), waits for a name $y$ on a channel with the same
name as the value of the local variable $n_A$ (the channel is specified by variable $n_A$)
and then stores $y$ in the local variable $m_A$.
The thread $Resp$ specifies the behavior of the responder.
Upon reception of a name $x$ on channel $c$, he stores it in the local variable
$n_B$, then creates a new name stored in local variable $m_B$ and finally
sends the value in $m_B$ on channel with the same  name as the value of $n_B$.
The thread $Main$ non-deterministically creates new thread instances of type $Init$ and $Resp$.
The local variable $x$ is used to store new names to be used for the creation of a new
thread instance.
Initially, all local variables of threads $Init/Resp$ are set to $\bot$.
In order to allow process instances to participate to several sessions
(potentially with different principals), we could also add the following rule
$$
\arrowup{stop_A}{restart}{init_{A}}\lguard n_A:=\bot,m_A:=\bot\rguard
$$
In this rule we require that {\em roles} and {\em identities} do not change from
session to session.\footnote{By means of thread and fresh name creation it is also possible
to specify a restart rule in which  a given process  takes a potential different
role or identity.}
\begin{figure*}[t]
$$
\begin{array}{l}
\begin{array}{l}
Thread~Init(local~id_A,n_A,m_A);
\smallskip\\
\begin{array}{ll}
~~~\arrowup{init_A}{fresh}{gen_A} & \lguard n_A:=new\rguard
\\
~~~\arrowup{gen_A}{c!\tuple{n_A}}{wait_A}&\lguard true\rguard
\\
~~~\arrowup{wait_A}{n_A?\tuple{y}}{stop_A}&\lguard m_A:=y\rguard
\end{array}\end{array}
\\
\\
\begin{array}{l}
Thread~Resp(local~id,n_B,m_B);\smallskip\\
\begin{array}{ll}
~~~\arrowup{init_B}{c?\tuple{x}}{gen_B}&\lguard n_B:=x\rguard
\\
~~~\arrowup{gen_B}{fresh}{ready_B}&\lguard m_B:={new}\rguard
\\
~~~\arrowup{ready_B}{n_B!\tuple{m_B}}{stop_B}&\lguard true\rguard
\end{array}\end{array}
\\
\\
\begin{array}{l}
Thread~Main(local~x);\smallskip\\
\begin{array}{ll}

~~~\arrowup{init_M}{id}{create}&\lguard x:=new\rguard
\\
~~~\arrowup{create}{new_A}{init_M}&\lguard run~Init~with~id_A:=x,n_A:=\bot,m_A:=\bot,x:=\bot\rguard
\\
~~~\arrowup{create}{new_B}{init_M}&\lguard run~Resp~with~id_B:=x,n_B:=\bot,m_B:=\bot,x:=\bot_B\rguard
\end{array}\end{array}
\end{array}
$$
\caption{Example of thread definitions.}
\label{model}
\end{figure*}
Starting from $G_0=\tuple{N_0,\tuple{init,\bot}}$, and running the {\em Main}
thread we can generate any number of copies of the threads $Init$
and $Resp$ each one with a unique identifier. Thus, we obtain
global configurations like
$$
\begin{array}{ll}
\langle N,&\tuple{init_M,\bot},\\
          &\tuple{init_A,i_1,\bot,\bot},\ldots,\tuple{init_A,i_K,\bot,\bot},\\
          & \tuple{init_B,i_{K+1},\bot,\bot},\ldots,\tuple{init_B,i_{K+L},\bot,\bot}~\rangle
\end{array}
$$
where $N=\{\bot,i_1,\ldots,i_K,i_{K+1},\ldots,i_{K+L}\}$ for $K,L\geq 0$. The
threads of type $Init$ and $Resp$ can start parallel sessions whenever
created. For $K=1$ and $L=1$ one possible session is as follows.
\\
Starting from
$$
\tuple{\{\bot,i_1,i_2\},\tuple{init_M,\bot},\tuple{init_A,i_1,\bot,\bot},\tuple{init_B,i_2,\bot,\bot}}\\
$$
if we apply the first rule of thread $Init$ to $\tuple{init_A,i_1,\bot,\bot}$ we obtain
$$
\tuple{\{\bot,i_1,i_2,a^1\},\tuple{init_M,\bot},\tuple{gen_A,i_1,a^1,\bot},\tuple{init_B,i_2,\bot,\bot}}
$$
where $a^1$ is the generated name ($a^1$ is distinct from $\bot$, $i_1$, and $i_2$).
Now if we apply the second rule of thread $Init$ and the first rule of thread $Resp$
(synchronization on channel $c$)
we obtain
$$
\tuple{\{\bot,i_1,i_2,a^1\},\tuple{init_M,\bot},\tuple{wait_A,i_1,a^1,\bot},\tuple{gen_B,i_2,a^1,\bot}}
$$
If we apply the second rule of thread $Resp$ we obtain
$$
\tuple{\{\bot,i_1,i_2,a^1,a^2\},\tuple{init_M,\bot},\tuple{wait_A,i_1,a^1,\bot},\tuple{ready_B,i_2,a^1,a^2}}
$$
Finally, if we apply the last rule of thread $Init$ and $Resp$
(synchronization on channel  $a^1$) we obtain
$$
\tuple{\{\bot,i_1,i_2,a^1,a^2\},\tuple{init_M,\bot},\tuple{stop_A,i_1,a^1,a^2},\tuple{stop_B,i_2,a^1,a^2}}
$$
Thus, at the end of the session the thread instances $i_1$ and $i_2$ have
both a local copy of the fresh names $a^1$ and $a^2$.
Note that a copy of the main thread $\tuple{init_M,\bot}$ is
always active in any reachable configuration, and, at any time, it
may  introduce new threads (either of type $Init$ or $Resp$) with
fresh identifiers. Generation of fresh names is also used by the
threads of type $Init$ and $Resp$ to create nonces. Furthermore,
threads can restart their life cycle (without changing
identifiers). Thus, in this example the set of possible reachable
configurations is infinite and contains configurations with {
arbitrarily many threads} and {fresh names}.
Since names are stored in the local variables of active threads,
the local data also range over an infinite domain.
\end{example}
\subsection{Expressive Power of TDL}
To study the expressive power of the TDL language, we will compare it
with the Turing equivalent formalism called {Two Counter Machines}.
A { Two Counters Machine} configurations is a tuple
$\tuple{\ell,c_1=n_1,c_2=n_2}$ where $\ell$ is control location taken from a
finite set $Q$,
and $n_1$ and $n_2$ are natural numbers that
represent the values of the counters $c_1$ and $c_2$.
Each counter can be incremented or decremented (if greater than zero)
by one.
Transitions combine operations on individual counters with changes
of control locations.
Specifically, the instructions for counter $c_i$ are as follows
$$
\begin{tabular}{l}
Inc: $\ell_1$: $c_i:=c_i+1$; goto $\ell_2$;\\
Dec: $\ell_1$: if $c_i>0$ then $c_i:=c_i-1$; goto $\ell_2$; else goto $\ell_3$;
\end{tabular}
$$
A { Two Counter Machine} consists then of a list of instructions and of
the initial state $\tuple{\ell_0,c_1=0,c_2=0}$.
The operational semantics is defined according to the intuitive semantics
of the instructions.
Problems like control state reachability are undecidable
for this computational model.
\\
The following property then holds.
\begin{theorem}
\label{twocounters}\rm
TDL programs can simulate { Two Counter Machines}.
\end{theorem}
\begin{proof}
In order to define a TDL program that simulates a
{ Two Counter Machine} we proceed as follows.
\begin{figure}[t]
$$
\begin{array}{l}
\mathrm{Thread}~Last(local~id,last,aux);\\
\begin{array}{l}
\begin{array}{l}
\smallskip\\
\mathbf{(Zero~test)}
\smallskip\\
\begin{array}{ll}
~~\arrowup{Idle}{Zero?\tuple{x}}{Busy} & \lguard id=x\rguard
\medskip\\
~~\arrowup{Busy}{tstC!\tuple{id,last}}{Wait}
\medskip\\
~~\arrowup{Wait}{nz?\tuple{x}}{AckNZ}& \lguard id=x\rguard
\medskip\\
~~\arrowup{Wait}{z?\tuple{x}}{AckZ}& \lguard id=x\rguard
\medskip\\
~~\arrowup{AckZ}{Yes!\tuple{id}}{Idle}
\medskip\\
~~\arrowup{AckNZ}{No!\tuple{id}}{idle}
\end{array}\end{array}
\\
\\
\begin{array}{l}
\mathbf{(Decrement)}
\\
\\
\begin{array}{ll}
~~\arrowup{Idle}{Dec?\tuple{x}}{Dbusy} & \lguard id=x\rguard
\\
\\
~~\arrowup{DBusy}{decC!\tuple{id,last}}{DWait}
\\
\\
~~\arrowup{DWait}{dack?\tuple{x,u}}{DAck} & \lguard id=x,last:=u\rguard
\\
\\
~~\arrowup{DAck}{DAck!\tuple{id}}{Idle}
\end{array}\end{array}
\\
\\
\begin{array}{l}
\mathbf{(Increment)}
\\
\\\begin{array}{ll}
~~\arrowup{Idle}{Inc?\tuple{x}}{INew}& \lguard id=x\rguard
\\
\\
~~\arrowup{INew}{new}{IRun}&\lguard aux:=new\rguard
\\
\\
~~\arrowup{IRun}{run}{IAck}&\lguard run~Cell~with~ idc:=id; prev:=last;next:=aux\rguard
\\
\\
~~\arrowup{IAck}{IAck!\tuple{id}}{Idle}&\lguard last:=aux\rguard
\end{array}\end{array}
\end{array}
\end{array}
$$
\caption{The process defining the last cell of the linked list
associated to a counter} \label{lastcell}
\end{figure}
Every counter is represented via a {\em doubly linked list} implemented via a
collection of threads of type $Cell$ and with a unique thread of type
$Last$ pointing to the head of the list.
\begin{figure}[t]
$$
\begin{array}{l}
\mathrm{Thread}~Cell(local~idc,prev,next);
\medskip\\
\begin{array}{l}
\mathbf{(Zero~test)}
\smallskip\\
~~\arrowup{idle}{tstC?\tuple{x,u}}{ackZ}~~\lguard x= idc,u=next,prev=next\rguard
\medskip\\
~~\arrowup{idle}{tstC?\tuple{x,u}}{ackNZ}~~\lguard  x=idc,u=next,prev\neq next \rguard
\medskip\\
~~\arrowup{ackZ}{z!\tuple{idc}}{idle}
\medskip\\
~~\arrowup{ackNZ}{nz!\tuple{idc}}{idle}
\end{array}
\\
\\
\begin{array}{l}
\mathbf{(Decrement)}
\\
\\
~~\arrowup{idle}{dec?\tuple{x,u}}{dec}~~\lguard x=idc,u=next,prev\neq next\rguard
\medskip\\
~~\arrowup{dec}{dack!\tuple{idc,prev}}{idle}
\end{array}
\end{array}
$$
\caption{The process defining a cell of the linked list associated
to a counter}
\label{thecell}
\end{figure}
The $i$-th counter having value zero is represented as the {\em empty
list} $Cell(i,v,v),Last(i,v,w)$ for some name $v$ and $w$ (we will explain later the use of
$w$).
The $i$-th counter having value $k$ is represented as
$$Cell(i,v_0,v_0),Cell(i,v_0,v_1),\ldots,C(i,v_{k-1},v_k),Last(i,v_k,w)$$
for distinct names $v_0,v_1,\ldots,v_k$.
The instructions on a counter are simulated by sending messages to the corresponding
$Last$ thread.
The messages are sent on channel $Zero$ (zero test), $Dec$ (decrement),
and $Inc$ (increment).
In reply to each of these messages, the thread $Last$ sends an acknowledgment, namely
$Yes/No$ for the zero test, $DAck$ for the decrement, $IAck$ for the increment operation.
$Last$ interacts with the $Cell$ threads via the messages  $tstC$, $decC$, $incC$
acknowledged by messages $z/nz$, $dack$. $iack$.
The interactions between a $Last$ thread and the $Cell$ threads is as follows.
\paragraph{Zero Test}
Upon reception of a message $\tuple{x}$ on channel $Zero$, the
$Last$ thread with local variables $id,last,aux$ checks that its identifier $id$
matches $x$ - see transition from $Idle$ to $Busy$ -
sends a message $\tuple{id,last}$ on channel $tstC$
directed to the cell pointed to by $last$ (transition from $Busy$ to $Wait$),
and then waits for an answer.
If the answer is sent on channel $nz$, standing for non-zero, (resp. $z$ standing for
zero) -  see transition from $Wait$ to $AckNZ$ (resp. $AckZ$) -
then it sends its identifier on channel $No$ (resp. $Yes$) as
an acknowledgment to the first message -
see transition from $AckNZ$ (resp. $Z$) to $Idle$.
As shown in Fig. \ref{thecell}, the thread $Cell$ with local variables
$idc$, $prev$, and $next$  that receives the message $tstC$, i.e.,
pointed to by a thread $Last$ with the same identifier as $idc$,
sends an acknowledgment on channel $z$ (zero) if $prev=next$, and on channel
$nz$ (non-zero) if  $prev\neq next$.
\paragraph{Decrement}
Upon reception of a message $\tuple{x}$ on channel $Dec$, the
$Last$ thread with local variables $id,last,aux$ checks
that its identifier $id$ matches $x$  (transition from $Idle$ to $Dbusy$),
sends a message $\tuple{id,last}$ on channel $decC$
directed to the cell pointed to by $last$ (transition from $Busy$ to $Wait$),
and then waits for an answer.
If the answer is sent on channel $dack$ (transition from $DWait$ to $DAck$)
then it updates the local variable $last$ with the pointer $u$ sent by the thread
$Cell$,
namely the $prev$ pointer of the cell pointed to by the current value of
$last$, and then  sends its identifier on channel $DAck$
to acknowledge the first message (transition from $DAck$ to $Idle$).

As shown in Fig. \ref{thecell}, a thread $Cell$ with local variables $idc$,
$prev$, and $next$   that receives the message $decC$ and
such that $next=last$ sends as an acknowledgment on channel  $dack$ the value $prev$.
\paragraph{Increment}
To simulate the increment operation, {\em Last} does not have to interact
with existing $Cell$ threads. Indeed, it only has to link a
new {\em Cell} thread to the head of the list (this is way the $Cell$ thread
has no operations to handle the increment operation).
As shown in Fig. \ref{lastcell} this can be done by creating a
new name stored in the local variable $aux$ (transition from $INew$ to $IRun$) and
spawning a new {\em Cell} thread (transition from $IRun$ to $IAck$)
with $prev$ pointer equal to $last$, and $next$ pointer equal to $aux$.
Finally, it acknowledges the increment request by sending its identifier
on channel $IAck$ and updates variable $last$ with the current value of $aux$.
\begin{figure}[t]
$$
\begin{array}{l}
\mathrm{Thread}~CM(local~id_1,id_2);
\medskip\\
~~~~~~~~~~\vdots
\medskip\\
\begin{array}{l}
\mathbf{(Instruction: \ell_1:~c_{i}:=c_{i}+1;~goto ~\ell_2;)}
\smallskip\\
~~\arrowup{\ell_1}{Inc!\tuple{id_i}}{wait_{\ell_1}}
\medskip\\
~~\arrowup{wait_{\ell_1}}{IAck!\tuple{x}}{\ell_2}~~\lguard x=id_i\rguard
\end{array}
\\
~~~~~~~~~~\vdots
\\
\begin{array}{l}
\mathbf{(Instruction: \ell_1:~c_{i}>0~then~c_{i}:=c_{i}-1;~goto ~\ell_2; else~goto~\ell_3;)}
\smallskip\\
~~\arrowup{\ell_1}{Zero!\tuple{id_i}}{wait_{\ell_1}}
\medskip\\
~~\arrowup{wait_{\ell_1}}{NZAck?\tuple{x}}{dec_{\ell_1}}~~\lguard x=id_i\rguard
\medskip\\
~~\arrowup{dec_{\ell_1}}{Dec!\tuple{id_i}}{wdec_{\ell_1}}
\medskip\\
~~\arrowup{wdec_{\ell_1}}{DAck?\tuple{y}}{\ell_2}~~\lguard y=id_i\rguard
\medskip\\
~~\arrowup{wait_{\ell_1}}{ZAck?\tuple{x}}{\ell_3}~~\lguard x=id_i\rguard
\end{array}
\\
~~~~~~~~~~\vdots
\end{array}
$$
\caption{The thread associated to a 2CM.}
\label{twocm}
\end{figure}

\begin{figure}[t]
$$
\begin{array}{l}
\mathrm{Thread}~Init(local~nid_1,p_1,nid_2,p_2);
\medskip\\
\begin{array}{c}
\begin{array}{l}
~~\arrowup{init}{freshId}{init_1}~~\lguard nid_1:=new\rguard
\medskip\\
~~\arrowup{init_1}{freshP}{init_2}~~\lguard p_1:=new\rguard
\medskip\\
~~\arrowup{init_2}{runC}{init_3}~~\lguard run~Cell~with~ idc:=nid_1; prev:=p_1;next:=p_1 \rguard
\medskip\\
~~\arrowup{init_3}{runL}{init_4}~~\lguard run~Last~with~ idc:=nid_1; last:=p_1;aux:=\bot \rguard
\medskip\\
~~\arrowup{init_4}{freshId}{init_5}~~\lguard nid_2:=new\rguard
\medskip\\
~~\arrowup{init_5}{freshP}{init_6}~~\lguard p_2:=new\rguard
\medskip\\
~~\arrowup{init_6}{runC}{init_7}~~\lguard run~Cell~with~ idc:=nid_2; prev:=p_2;next:=p_2\rguard
\medskip\\
~~\arrowup{init_7}{runL}{init_8}~~\lguard run~Last~with~ idc:=nid_2; last:=p_2;aux:=\bot \rguard
\medskip\\
~~\arrowup{init_8}{runCM}{init_9}~~\lguard run~2CM~with~ id_1:=nid_1; id_2:=nid_2\rguard
\end{array}
\end{array}
\end{array}
$$
\caption{The initialization thread.}
\label{init}
\end{figure}
\paragraph{Two Counter Machine Instructions}
We are now ready to use the operations provided by the thread $Last$
to simulate the instructions of a Two Counter Machine.
As shown in Fig. \ref{twocm}, we use a thread $CM$ with two local variables
$id_1,id_2$ to represent the list of instructions of a 2CM with counters $c_1,c_2$.
Control locations of the Two Counter Machines are used as local states of the thread
$CM$. The initial local state of the $CM$ thread is the initial control location.
The increment instruction on counter $c_i$ at control location $\ell_1$ is simulated by an
handshaking with the $Last$ thread with identifier $id_i$: we first send the message
$Inc!\tuple{id_i}$, wait for the acknowledgment on channel $IAck$ and then move to state $\ell_2$.
Similarly, for the decrement instruction on counter $c_i$ at
control location $\ell_1$
we first send the message $Zero!\tuple{id_i}$. If we receive an acknowledgment
on channel $NZAck$  we send a $Dec$ request, wait for completion
and then move to $\ell_2$.
If we receive an acknowledgment on channel $ZAck$  we directly move to $\ell_3$.
\paragraph{Initialization}
The last step of the encoding is the definition of the initial state of the system.
For this purpose, we use the thread $Init$ of Fig. \ref{init}.
The first four rules of $Init$ initialize the first counter:
they create two new names $nid_1$ (an identifier for counter $c_1$)
and $p_1$, and then spawn the new threads $Cell(nid_1,p_1,p_1),Last(nid_1,p_1,\bot)$.
The following four rules spawns the new threads
$Cell(nid_2,p_2,p_2),Last(nid_2,p_2,\bot)$.
After this stage, we create a thread of type $2CM$ to start the simulation of the
instructions of the Two Counter Machines.
The initial configuration of the whole system is $G_0=\tuple{init,\bot,\bot}$.
By construction we have that an execution step from
$\tuple{\ell_1,c_1=n_1,c_2=n_2}$ to $\tuple{\ell_2,c_1=m_1,c_2=m_2}$
is simulated by an execution run going from a global configuration in which
the local state of thread $CM$ is $\tuple{\ell_1,id_1,id_2}$ and in which
we have $n_i$ occurrences of thread $Cell$ with the same identifier $id_i$ for $i:1,2$,
to a global configuration in which
the local state of thread $CM$ is $\tuple{\ell_2,id_1,id_2}$ and in which
we have $m_i$ occurrences of thread $Cell$ with the same identifier $id_i$ for $i:1,2$.
Thus, every executions of a 2CM $M$ corresponds to an execution
of the corresponding TDL program that starts from the initial configuration
$G_0=\tuple{init,\bot,\bot}$.
\end{proof}
As a consequence of the previous
theorem, we have the following corollary.
\begin{corollary}\rm
\label{reachability} Given a TDL program, a global configurations
$G$, and a control location $\ell$, deciding if  there exists a run
going from $G_0$ to a global configuration that contains $\ell$
(control state reachability) is an undecidable problem.
\end{corollary}
\section{From TDL to MSR$_{NC}$}
\label{MSRTranslation}
As mentioned in the introduction, our
verification methodology is based on a translation of TDL programs
into low level specifications given in MSR$_{NC}$. Our goal is to
extend the connection between CCS and Petri Nets \cite{GS92} to
TDL and MSR so as to be able to apply the verification methods
defined in \cite{Del02} to multithreaded programs. In
the next section we will summarize the main features of the
language MSR$_{NC}$ introduced in \cite{Del01}.
\subsection{Preliminaries on MSR$_{NC}$}
\label{MSRlanguage}
 $NC$-constraints are { linear arithmetic constraints} in
which conjuncts have one of the following form: $true$, $x=y$,
$x>y$, $x=c$, or $x>c$, $x$ and $y$ being two variables from a
denumerable set $\calV$ that range over the rationals, and $c$
being an integer. The {\em solutions} $Sol$ of a constraint
$\varphi$ are defined as all evaluations (from $\calV$ to $\Rat$)
that satisfy $\varphi$. A {\em constraint} $\varphi$ is {\em
satisfiable} whenever $Sol(\varphi)\neq\emptyset$. Furthermore,
$\psi$ {\em entails} $\varphi$ whenever $Sol(\psi)\subseteq
Sol(\varphi)$.  $NC$-constraints are closed under elimination of
existentially quantified variables.
\smallskip\\
Let $\calP$ be a set of predicate symbols. An {\em atomic formula}
$p(x_1,\ldots,x_n)$ is such that $p\in\calP$, and $x_1,\ldots,x_n$
are {\em distinct} variables in $\calV$. A {\em multiset} of
atomic formulas is indicated as $A_1~|~\ldots~|~A_k$, where $A_i$
and $A_j$ have distinct variables (we  use variable renaming if necessary),
and $|$ is the multiset constructor.
\smallskip\\
In the rest of the paper we will use $\calM$, $\calN$,
$\ldots$ to denote {\em multisets} of atomic formulas,
$\epsilon$ to denote the {\em empty multiset},
$\oplus$ to denote {\em multiset union}
and $\ominus$ to denote {\em multiset difference}.
An MSR$_{NC}$ {\em configuration}
is a multiset of {\em ground atomic formulas}, i.e., atomic
formulas like $p(d_1,\ldots,d_n)$ where $d_i$ is a rational for
$i:1,\ldots,n$.
\smallskip\\
An MSR$_{NC}$ {\em rule} has the form
$\calM~\longrightarrow\calM'~{:}~\varphi$, where $\calM$ and
$\calM'$ are two (possibly empty) multisets of atomic formulas
 with {\em distinct} variables built on predicates in $\calP$, and
$\varphi$ is an $NC$-constraint. The ground instances of an
MSR$_{NC}$ rule are defined as
$$
Inst(\calM\longrightarrow\calM':\varphi)=\{
\sigma(\calM)\longrightarrow\sigma(\calM')~|~\sigma~\in~Sol(\varphi)\}
$$
where $\sigma$ is extended in the natural way to multisets,
i.e., $\sigma(\calM)$ and $\sigma(\calM')$ are MSR$_{NC}$ configurations.
\smallskip\\
An MSR$_{NC}$ {\em specification} $\calS$ is a tuple
$\tuple{\calP,\calI,\calR}$, where $\calP$ is a finite set of predicate
symbols, $\calI$ is finite a set of ({\em initial}) MSR$_{NC}$
configurations, and $\calR$ is a finite set of MSR$_{NC}$ rules over
$\calP$.
\smallskip\\
The operational semantics describes the update from
a configuration $\calM$ to one of its possible successor configurations $\calM'$.
$\calM'$ is obtained from $\calM$ by rewriting (modulo associativity and commutativity)
the left-hand side
of an instance of a rule into the corresponding right-hand side.
In order to be fireable, the left-hand side must be included in $\calM$.
Since instances and rules are selected in a non deterministic way, in general a configuration
can have a (possibly infinite) set of (one-step) successors.\\
\smallskip\\
Formally, a rule $\calH~{\longrightarrow}~\calB~:~\varphi$ from $\calR$
is  enabled at $\calM$ {\em via } the ground substitution $\sigma\in Sol(\varphi)$
if and only if $\sigma(\calH)\preccurlyeq \calM$.
Firing rule $R$ enabled at $\calM$ via $\sigma$ yields the new configuration
$$
\calM'~=~\sigma(\calB) \oplus (\calM\ominus \sigma(\calH))
$$
We use $\calM\Rightarrow_{MSR} \calM'$ to denote the firing of a rule at $\calM$
yielding $\calM'$.
\smallskip\\
A run is a sequence of configurations
$\calM_0\calM_1\ldots\calM_k$ with $\calM_0\in\calI$ such that
$\calM_{i}\Rightarrow_{MSR}\calM_{i+1}$ for $i\geq 0$.
A configuration $\calM$ is reachable if there exists $\calM_0\in\calI$ such that
$\calM_0\stackrel{*}{\Rightarrow}_{MSR}\calM$, where
$\stackrel{*}{\Rightarrow_{MSR}}$ is the transitive closure of $\Rightarrow_{MSR}$.
Finally, the successor and predecessor operators $Post$ and $Pre$ are defined
on a set of configurations $S$ as
$Post(S)=\{\calM'|\calM\Rightarrow_{MSR}\calM',\calM\in S\}
$ and $Pre(S)=\{\calM|\calM\Rightarrow_{MSR}\calM',\calM'\in S\}$, respectively.
$Pre^*$ and $Post^*$ denote their transitive closure.
\comment{
\begin{example}
The MSR$_{NC}$ rule $p(x)~|~q(y)\rightarrow
s(x')~|~p(y'):x=y,y'>x',x'>y$ symbolically represents a family of
rewriting rules on ground atomic formulas obtained instantiating
the variables with solution of the constraint. The ground multiset
rewriting rule $p(1)~|~q(1)\rightarrow s(3)~|~p(5)$ is one of its
instances. Given the instance $p(1)~|~q(1)\rightarrow s(3)~|~p(5)$
of the rule $p(x)~|~q(y)\rightarrow_{MSR}
s(x')~|~p(y'):x=y,y'>x',x'>y$, a possible rewriting step is
 $p(3)~|~p(1)~|~q(1)~|~r(4)\Rightarrow p(3)~|~s(3)~|~p(5)~|~r(4)$.
\end{example}}
\smallskip\\
 As shown in \cite{Del01,BD02}, Petri Nets represent a natural
 abstractions of MSR$_{NC}$ (and more in general of MSR rule with constraints) specifications.
 They can be encoded, in fact, in {\em propositional} MSR specifications
 (e.g. abstracting away arguments from atomic formulas).
\subsection{Translation from TDL to MSR$_{NC}$}
The first thing to do is to find an adequate representation of names.
Since all we need is a way to distinguish old and  new names,
we just need an infinite domain in which the $=$ and $\neq$ relation
are supported. Thus, we can interpret names in $\calN$ either as
integer of as rational numbers.
Since operations like variable elimination are computationally less expensive
than over integers, we choose to view names as non-negative rationals.
Thus, a local (TDL) configuration $p=\tuple{s,n_1,\ldots,n_k}$ is
encoded as the atomic formula $\TDLtoMSR{p}=s(n_1,\ldots,n_k)$,
where $n_i$ is a non-negative rational.
Furthermore, a global (TDL) configuration
$G=\tuple{N,p_1,\ldots,p_m}$ is encoded as an MSR$_{NC}$
configuration $\TDLtoMSR{G}$
$$
\TDLtoMSR{p_1}~|~\ldots~|~\TDLtoMSR{p_m}~|~fresh(n)
$$
where the value $n$ in the auxiliary atomic formula $fresh(n)$ is
an rational number strictly greater than all values occurring in
$\TDLtoMSR{p_1},\ldots,\TDLtoMSR{p_m}$.
The predicate $fresh$ will allow us to generate unused names every time needed.
\smallskip\\
The translation of constants $\calC=\{c_1,\ldots,c_m\}$,
and variables is defined as follows:
$\TDLtoMSR{x}=x$ for $x\in\calV$, $\TDLtoMSR{\bot}=0$,
$\TDLtoMSR{c_i}=i$ for $i:1,\ldots,m$.
We extend ~$\TDLtoMSR{\cdot}$~ in the natural way on a guard $\gamma$, by
decomposing every formula $x\neq e$ into $x<\TDLtoMSR{e}$ and
$x>\TDLtoMSR{e}$. We will call $\TDLtoMSR{\gamma}$ the resulting
{\em set} of $NC$-constraints.\footnote{As an example, if $\gamma$
is the constraint $x=1,x\neq z$ then $\TDLtoMSR{\gamma}$ consists
of the two constraints $x=1,x>z$ and $x=1,z>x$.}
\smallskip\\
Given $V=\{x_1,\ldots,x_k\}$, we define $V'$ as the set of new
variables $\{x_1',\ldots,x_k'\}$. Now, let us consider the
assignment $\alpha$ defined as $x_1:=e_1,\ldots,x_k:=e_k$ (we add
assignments like $x_i:=x_i$ if some variable does not occur as
target of $\alpha$). Then,  $\TDLtoMSR{\alpha}$ is the
$NC$-constraint $x_1'=\TDLtoMSR{e_1},\ldots,x_k'=\TDLtoMSR{e_k}$.
\smallskip\\
The translation of thread definitions is defined below (where we
will often refer to Example \ref{AliceBob}).
\paragraph{Initial Global Configuration}
Given an initial global configuration consisting of the local configurations
 $\tuple{s_i,n_{i1},\ldots,n_{ik_i}}$ with $n_{ij}=\bot$ for $i:1,\ldots,u$,
 we define the following MSR$_{NC}$ rule
$$
\begin{array}{l}
init\rightarrow s_1(x_{11},\ldots,x_{1k1})~|~\ldots~|~
s_u(x_{u1},\ldots,x_{uk_u})~|~fresh(x)~:~\\
~~~~~~~~~~~~~~~~~~~~x>C,x_{11}=0,\ldots,x_{uk_u}=0
\end{array}
$$
here $C$ is the largest rational used to interpret the constants in
$\calC$.
\smallskip\\
For each thread definition $P=\tuple{Q,s_0,V,R}$ in $\calT$ with
$V=\{x_1,\ldots,x_k\}$ we translate the rules in $R$ as described
below.
\paragraph{Internal Moves}
For every {\em internal move} $\arrowup{s}{a}{s'}\lguard \gamma,\alpha\rguard$, and
every $\nu\in\TDLtoMSR{\gamma}$ we define
$$
s(x_1,\ldots,x_k)\rightarrow s'(x_1',\ldots,x_k'):\nu,\TDLtoMSR{\alpha}
$$
\paragraph{Name Generation}
For every {\em name generation} $\arrowup{s}{a}{s'}\lguard x_i:=new\rguard$,
we define
$$
s(x_1,\ldots,x_k)~|~fresh(x)\rightarrow
s'(x_1',\ldots,x_k')~|~fresh(y): y>x'_i,x_i'>x,\bigwedge_{j\neq i} x_j'=x_j
$$
For instance, the name generation
$\arrowup{init_A}{fresh}{gen_A}\lguard n:=new\rguard$ is mapped into the
MSR$_{NC}$ rule
$\arrowup{init_A(id,x,y)|~fresh(u)}{}{gen_A(id',x',y')~|~fresh(u')}:\varphi$
where $\varphi$ is the constraint $u'>x',x'>u,y'=y,id'=id$. The
constraint $x'>u$ represents the fact that the new name associated
to the local variable $n$ (the second argument of the atoms
representing the thread) is fresh, whereas $u'>x'$ updates the
current value of $fresh$ to ensure that the next generated names
will be picked up from unused values.
\paragraph{Thread Creation}
Let $P=\tuple{Q',t_0,V',R'}$ and $V'=\{y_{1},\ldots,y_{u}\}$.
Then, for every {\em thread creation} $\arrowup{s}{a}{s'}\lguard
run~P~with~\alpha\rguard$,
we define
$$
\begin{array}{l}
s(x_1,\ldots,x_k)\rightarrow
s'(x_1',\ldots,x_k')~|~t(y_{1}',\ldots,y_{u}')~:~
x_1'=x_1,\ldots,x_k'=x_k,\TDLtoMSR{\alpha}.
\end{array}
$$
E.g., consider the rule
$\arrowup{create}{new_A}{init_M}\lguard run~Init~with~id:=x,\ldots\rguard$ of Example
\ref{AliceBob}. Its encoding yields the MSR$_{NC}$ rule
$\arrowup{create(x)}{}{init_M(x')~|~init_{A}(id',n',m')}:\psi$, where
$\psi$ represents the initialization of the local variables of the new thread
$x'=x,id'=x,n'=0,m'=0$.
\paragraph{Rendez-vous}
The encoding of rendez-vous communication is based on the use of
constraint operations like variable elimination.
Let $P$ and $P'$ be a pair of thread definitions, with
local variables $V=\{x_1,\ldots,x_k\}$ and
$V'=\{y_1,\ldots,y_l\}$ with $V\cap V'=\emptyset$. We first select
all rules $\arrowup{s}{e!m}{s'}\lguard \gamma,\alpha\rguard$ in $R$ and
$\arrowup{t}{e'?m'}{t'}\lguard \gamma',\alpha'\rguard$ in $R'$, such that
$m=\tuple{w_1,\ldots,w_u}$, $m'=\tuple{w_1',\ldots,w_v'}$ and
$u=v$.
Then, we define the new MSR$_{NC}$ rule
$$
s(x_1,\ldots,x_k)~|~t(y_1,\ldots,y_l) \rightarrow
s'(x_1',\ldots,x_k')~|~t'(y_1',\ldots,y_l') :\varphi
$$
for every $\nu\in\TDLtoMSR{\gamma}$ and
$\nu'\in\TDLtoMSR{\gamma'}$ such that the NC-constraint $\varphi$
obtained by eliminating
$w_1',\ldots,w_v'$ from the constraint
$\nu\wedge\nu'\wedge\TDLtoMSR{\alpha}\wedge\TDLtoMSR{\alpha'}\wedge
w_1=w_1'\wedge\ldots\wedge w_v=w_v'$ is {\em satisfiable}.
For instance, consider the rules
$\arrowup{wait_A}{n_A?\tuple{y}}{stop_A}\lguard m_A:=y\rguard$ and
$\arrowup{ready_B}{n_B!\tuple{m_B}}{stop_B}\lguard true\rguard$.
We first build up a new constraint by conjoining the NC-constraints $y=m_B$
(matching of message templates), and
$n_A=n_B,m_A'=y,n_A'=n_A,m_B'=m_B,n_B'=n_B,id_1'=id_1,id_2'=id_2$
(guards and actions of sender and receiver).
After eliminating $y$ we obtain the constraint $\varphi$ defined as
$n_B=n_A,m_A'=m_B,n_A'=n_A,m_B'=m_B,n_B'=n_B,id_1'=id_1,id_2'=id_2$
defined over the variables of the two considered threads. This
step allows us to {\em symbolically} represent the passing of
names. After this step, we can represent the synchronization of
the two threads by using a rule that simultaneously rewrite all
instances that satisfy the constraints on the local data expressed
by $\varphi$, i.e., we obtain the rule
$$
\begin{array}{l}
wait_A(id_1,n_A,m_A)|~ready_B(id_2,n_B,m_B)\longrightarrow\\
~~~~~~~~~~~~~~stop_A(id_1',n_A',m_A')~|~stop_B(id_2',n_B',m_B'):\varphi
\end{array}
$$
\begin{figure*}[t]
{
$
\begin{array}{l}
\arrowup{init}{}{fresh(x)~|~init_M(y)}~:~x>0,y=0.
\smallskip\\
\arrowup{fresh(x)~|~init_{M}(y)}{}{fresh(x')~|~create(y')}~:~x'>y',y'>x.
\smallskip\\
\arrowup{create(x)}{}{init_M(x')~|~init_{A}(id',n',m')}~:~x'=x,id'=x,n'=0,m'=0.
\smallskip\\
\arrowup{create(x)}{}{init_M(x')~|~init_{B}(id',n',m')}~:~x'=x,id'=x,n'=0,m'=0.
\smallskip\\
\arrowup{init_A(id,n,m)|~fresh(u)}{}
        {gen_A(id,n',m)~|~fresh(u')}~:~u'>n',n'>u.
\smallskip\\
\arrowup{gen_A(id_1,n,m)|~init_B(id_2,u,v)}{}
        {wait_A(id_1,n,m)~|~gen_B(id_2',u',v')}~:~u'=n,v'=v
\smallskip\\
\arrowup{gen_B(id,n,m)|~fresh(u)}{}
        {ready_B(id,n,m')~|~fresh(u')}~:~u'>m',m'>u.
\smallskip\\
\arrowup{wait_A(id_1,n,m)|~ready_B(id_2,u,v)}{}
        {stop_A(id_1,n,m')~|~stop_B(id_2,u,v)}~:~
n=u,m'=v.
\smallskip\\
\arrowup{stop_A(id,n,m)}{}
        {init_A(id',n',m')}~:~n'=0,m'=0,id'=id.
\smallskip\\
\arrowup{stop_B(id,n,m)}{}
        {init_B(id',n',m')}~:~n'=0,m'=0,id'=id.
\end{array}
$
} \caption{Encoding of Example \ref{AliceBob}: for simplicity we embed constraints like $x=x'$
into the MSR formulas.} \label{ABMSR}
\end{figure*}
The complete translation of Example \ref{AliceBob} is shown in
Fig. \ref{ABMSR} (for simplicity we have applied a renaming of
variables in the resulting rules). An example of run in the
resulting MSR$_{NC}$ specification is shown in Figure \ref{runABMSR}.
\begin{figure*}[t]
{
$
\begin{array}{l}
init\Rightarrow\ldots\Rightarrow
fresh(4)~|~init_M(0)~|~init_A(2,0,0)~|~init_B(3,0,0)\\
\Rightarrow
fresh(8)~|~init_M(0)~|~gen_A(2,6,0)~|~init_B(3,0,0)\\
\Rightarrow
fresh(8)~|~init_M(0)~|~wait_A(2,6,0)~|~gen_B(3,6,0)\\
\Rightarrow\ldots\Rightarrow
fresh(16)~|~init_M(0)~|~wait_A(2,6,0)~|~gen_B(3,6,0)~|~init_A(11,0,0)
\end{array}
$}
\caption{A run in the encoded program.} \label{runABMSR}
\end{figure*}
Note that, a fresh name is selected between all values strictly
greater than the current value of $fresh$ (e.g. in the second step
$6>4$), and then $fresh$ is updated to a value strictly greater
than all newly generated names (e.g. $8>6>4$).

Let $\calT=\tuple{P_1,\ldots,P_t}$ be a collection of thread
definitions and $G_0$ be an initial global state. Let
$\calS$ be the MSR$_{NC}$ specification that results from the
translation described in the previous section.
\smallskip\\
Let $G=\tuple{N,p_1,\ldots,p_n}$ be a global configuration with
$p_i=\tuple{s_i,v_{i1},\ldots,v_{ik_i}}$, and let $h:N\leadsto \Rat_+$
be an injective  mapping. Then, we define $\TDLtoMSR{G}(h)$ as the
MSR$_{NC}$ configuration
$$
s_1(h(v_{11}),\ldots,h(v_{1k_1}))~|~\ldots~|~s_n(h(v_{n1}),\ldots,h(v_{nk_n}))~|~fresh(v)
$$
where $v$ is a the first value strictly greater than all values in the range of $h$.
Given an MSR$_{NC}$ configuration $\calM$ defined as
$s_1(v_{11},\ldots,v_{1k_1})~|~\ldots~|~s_n(v_{n1},\ldots,v_{nk_n})$
with $s_{ij}\in \Rat_+$, let $V(\calM)\subseteq\Rat_+$ be the set
of values occurring in $\calM$. Then,  given a bijective
mapping $f:V(\calM)\leadsto N\subseteq \calN$, we
define $\TDLtoMSR{\calM}(f)$ as the global configuration
$\tuple{N,p_1,\ldots,p_n}$ where
$p_i=\tuple{s_i,f(v_{i1}),\ldots,f(v_{ik_i})}$.
\smallskip\\
Based on the previous definitions, the following property then holds.
\begin{theorem}\rm
\label{soundcomp}
For every run $G_0 G_1 \ldots$ in $\calT$ with corresponding set
of names $N_0 N_1\ldots$, there exist sets $D_0 D_1 \ldots$ and
bijective mappings $h_0 h_1 \ldots$ with
$h_i:N_i\leadsto D_i\subseteq\Rat_+$ for $i\geq 0$, such that
$init~\TDLtoMSR{G_0}({h_0})\TDLtoMSR{G_1}({h_1})\ldots$ is a run of
$\calS$. Vice versa, if $init~\calM_0\calM_1\ldots$ is a run of
$\calS$, then there exist sets $N_0 N_1 \ldots$ in $\calN$ and
bijective mappings $f_0 f_1 \ldots$ with
$f_i:V(\calM_i)\leadsto N_i$ for $i\geq 0$, such that
$\TDLtoMSR{\calM_0}({f_0})\TDLtoMSR{\calM_1}({f_1})\ldots$ is a run in
$\calT$.
\end{theorem}

\begin{proof}
We first prove that every run in $\calT$ is simulated by a run in $\calS$.

Let $G_0\ldots G_l$ be a run in $\calT$, i.e., a sequence
of global states  (with associated set of names $N_0\ldots N_l$)
such that $G_i\Rightarrow G_{i+1}$ and $N_i\subseteq N_{i+1}$ for $i\geq 0$.
\\
We prove that it can be simulated in $\calS$ by induction on its length $l$.
\\
Specifically, suppose that there exist
 sets of non negative rationals $D_0 \ldots D_{l}$ and bijective mappings $h_0 \ldots h_{l}$ with
$h_i:N_i\leadsto D_i$ for $0\leq i \leq {l}$, such that

$$init~\widehat{G_0}({h_0})\ldots\widehat{G_l}({h_l})$$
is a run of $\calS$.
Furthermore, suppose $G_{l}\Rightarrow G_{l+1}$.
\\
We prove the thesis by  a case-analysis on the type
of rule applied in the last step of the run.
\\
Let $G_l=\tuple{N_l,p_1,\ldots,p_r}$ and
$p_j=\tuple{s,n_1,\ldots,n_k}$ be a local configuration for the
thread definition
$P=\tuple{Q,s,V,R}$ with $V=\{x_1,\ldots,x_k\}$
and $n_i\in N_l$ for $i:1,\ldots,k$.
\\
{\em Assignment} Suppose there exists a rule $\arrowup{s}{a}{s'}[\gamma,\alpha]$ in $R$
such that $\rho_{p_j}$ satisfies $\gamma$,
$G_l=\tuple{N_l,\ldots,p_j,\ldots}\Rightarrow\tuple{N_{l+1},\ldots,p_j',\ldots}=G_{l+1}$
$N_l=N_{l+1}$,
$p_j'=\tuple{s',n_1',\ldots,n_k'}$,
and if $x_i:=y_i$ occurs in $\alpha$, then $n_i'=\rho_{p_j}(y_i)$,
otherwise $n_i'=n_i$ for $i:1,\ldots,k$.
\\
The encoding of the rule returns one $MSR_{NC}$ rule having the form
$$
s(x_1,\ldots,x_k)\rightarrow s'(x_1',\ldots,x_k'):\gamma',\widehat{\alpha}
$$
for every $\gamma'\in\widehat{\gamma}$.
\\
By inductive hypothesis,
 $\widehat{G_l}({h_l})$ is a multiset of atomic formulas that contains the
formula $s(h_l(n_1),\ldots,h_l(n_k))$.
\\
Now let us define $h_{l+1}$ as the mapping from $N_l$ to $D_l$ such that
$h_{l+1}(n_i')=h_l(n_j)$ if $x_i:=x_j$ is in $\alpha$ and
$h_{l+1}(n_i')=0$ if $x_i:=\bot$ is in $\alpha$.
Furthermore, let us the define the evaluation
$$
\sigma=\tuple{x_1\mapsto h_l(n_1),\ldots,x_k\mapsto h_l(n_k),
        x_1'\mapsto h_{l+1}(n_1'),\ldots,x_k'\mapsto h_{l+1}(n_k')}
$$
\\
Then, by construction of  the set of constraints $\widehat{\gamma}$ and of
the constraint $\widehat{\alpha}$,
it follows that $\sigma$ is a solution for $\gamma',\widehat{\alpha}$ for some
$\gamma'\in\widehat{\gamma}$.
As a consequence, we have that
$$
s(n_1,\ldots,n_k)\rightarrow s'(n_1',\ldots,n_k')
$$
is a ground instance of one of the considered $MSR_{NC}$ rules.
\\
Thus, starting from the $MSR_{NC}$ configuration
$\widehat{G_l}({h_l})$, if we apply a rewriting step we obtain  a
new configuration in which $s(n_1,\ldots,n_k)$ is replaced by $s'(n_1',\ldots,n_k')$,
and all the other atomic formulas in $\widehat{G_{l+1}}({h_{l+1}})$ are the same as
in $\widehat{G_l}({h_l})$.
The resulting $MSR_{NC}$ configuration coincides then with the definition of
$\widehat{G_{l+1}}({h_{l+1}})$.
\\
{\em Creation of new names}
Let us now consider the case of fresh name generation.
Suppose there exists a rule $\arrowup{s}{a}{s'}[x_i:=new]$ in $R$,
and let $n\not\in N_l$,
and suppose
 $
\tuple{N_l,\ldots,p_j,\ldots}\Rightarrow\tuple{N_{l+1},\ldots,p_j',\ldots}
 $
where $N_{l+1}=N_l\cup\{v\}$,
 $p_j'=\tuple{s',n_1',\ldots,n_k'}$ where $n_i'=n$, and $n_j'=n_j$ for $j\neq i$.
\\
We note than that the encoding of the previous rule returns the $MSR_{NC}$ rule
$$
s(x_1,\ldots,x_k)~|~fresh(x)\rightarrow s'(x_1',\ldots,x_k')~|~fresh(x'):\varphi
$$
where $\varphi$ consists of the constraints  $y>x'_i,x_i'>x$ and $x_j'=x_j$ for $j\neq i$.
By inductive hypothesis,
 $\widehat{G_l}({h_l})$ is a multiset of atomic formulas that contains the
formulas $s(h_l(n_1),\ldots,h_l(n_k))$ and $fresh(v)$ where  $h_l$ is a mapping into $D_l$,
and $v$ is the first non-negative rational strictly greater than all
values occurring in the formulas denoting processes.
\\
Let $v$ be a non negative rational strictly greater than all values in $D_l$.
Furthermore, let us define $v'=v+1$ and
$D_{l+1}=D_l\cup\{v,v'\}$.
\\
Furthermore, we define $h_{l+1}$ as follows
$h_{l+1}(n)=h_{l}(n)$ for $n\in N_l$, and $h_{l+1}(n_i')=h_{l+1}(n)=v'$.
Furthermore, we define the following evaluation
$$
\begin{array}{lcl}
\sigma & = \langle &
                x\mapsto v,x_1\mapsto h_l(n_1),\ldots,x_k\mapsto h_l(n_k),\\
       & &      x'\mapsto v',x_1'\mapsto h_{l+1}(n_1'),\ldots,x_k'\mapsto h_{l+1}(n_k')
        ~~~\rangle
\end{array}
$$
Then, by construction of $\sigma$ and $\widehat{\alpha}$,
it follows that $\sigma$ is a solution for $\widehat{\alpha}$.
Thus,
$$
s(n_1,\ldots,n_k)~|~fresh(v)\rightarrow s'(n_1',\ldots,n_k')~|~fresh(v')
$$
is a ground instance of the considered $MSR_{NC}$ rule.
\\
Starting from the $MSR_{NC}$ configuration
$\widehat{G_l}({h_l})$, if we apply a rewriting step we obtain  a
new configuration in which $s(n_1,\ldots,n_k)$ and $fresh(v)$ are
substituted by $s'(n_1',\ldots,n_k')$ and $fresh(v')$,
and all the other atomic formulas in $\widehat{G_{l+1}}({h_{l+1}})$ are the same
as in $\widehat{G_l}({h_l})$.
We conclude by noting that this formula coincides with the definition
of $\widehat{G_{l+1}}({h_{l+1}})$.
\\
For sake of brevity we omit the case of {\em thread creation}
whose only difference from the previous cases is the creation of
several new atoms instead (with values obtained by evaluating the action)
of only one.
\\
{\em Rendez-vous}
Let
$p_i=\tuple{s,n_1,\ldots,n_k}$ and $p_j=\tuple{t,m_1,\ldots,m_u}$
two local configurations for threads  $P\neq P'$,
$n_i\in N_l$ for $i:1,\ldots,k$ and $m_i\in N_l$ for
$i:1,\ldots,u$.
\\
Suppose
$\arrowup{s}{c!m}{s'}[\gamma,\alpha]$ and
$\arrowup{t}{c?m'}{t'}[\gamma',\alpha']$,
where $m=\tuple{x_{i_1},\ldots,x_{i_v}}$, and
$m'=\tuple{y_1,\ldots,y_v}$ (
all defined over distinct variables)
are rules in $R$.
\\
Furthermore, suppose that $\rho_{p_i}$ satisfies $\gamma$, and that $\rho'$
(see definition of the operational semantics)
satisfies $\gamma'$, and suppose that
 $G_l=\tuple{N_l,\ldots,p_i,\ldots,p_j,\ldots}\Rightarrow
\tuple{N_{l+1},\ldots,p_i',\ldots,p_j',\ldots}=G_{l+1},
 $
where $N_{l+1}=N_l$, $p_i'=\tuple{s',n_1',\ldots,n_k'}$,
$p_j'=\tuple{t',m_1',\ldots,m_u'}$,
and if $x_i:=e$ occurs in $\alpha$, then $n_i'=\rho_{p_i}(e)$,
otherwise $n_i'=n_i$ for $i:1,\ldots,k$;
if $u_i:=e$ occurs in $\alpha'$, then $m_i'=\rho'(e)$,
otherwise $m_i'=m_i$ for $i:1,\ldots,u$.
\\
By inductive hypothesis,
 $\widehat{G_l}({h_l})$ is a multiset of atomic formulas that contains the
formulas $s(h_l(n_1),\ldots,h_l(n_k))$ and $t(h_l(m_1),\ldots,h_l(m_u))$.
\\
Now, let us define $h_{l+1}$ as the mapping from $N_l$ to $D_l$ such that
 $h_{l+1}(n_i')=h_l(n_j)$ if $x_i:=x_j$ is in $\alpha$,
$h_{l+1}(m_i')=h_l(m_j)$ if $u_i:=u_j$ is in $\alpha'$,
 $h_{l+1}(n_i')=0$ if $x_i:=\bot$ is in $\alpha$,
 $h_{l+1}(m_i')=0$ if $u_i:=\bot$ is in $\alpha'$.
 \\
Now, let us define $\sigma$ as the evaluation from $N_l$ to $D_l$ such that
$$
\begin{array}{l}
\sigma=\sigma_1\cup\sigma_2\\
\sigma_1=\tuple{x_1\mapsto h_l(n_1),\ldots,x_k\mapsto h_l(n_k),
               u_1\mapsto h_l(m_1),\ldots,u_u\mapsto h_l(m_u)}
\\
\sigma_2=\tuple{x_1'\mapsto h_{l+1}(n_1'),\ldots,x_k'\mapsto h_{l+1}(n_k'),
            u_1'\mapsto h_{l+1}(m_1'),\ldots,u_u'\mapsto h_{l+1}(m_u')}.
\end{array}
$$
Then, by construction of
the sets of constraints $\widehat{\gamma},\widehat{\gamma'},\widehat{\alpha}$ and
$\widehat{\alpha'}$ it follows that $\sigma$ is a solution for the constraint
$\exists w_1'.\ldots\exists w_p'.
\theta\wedge\theta'\wedge\widehat{\alpha}\wedge\widehat{\alpha'}\wedge
w_1=w_1'\wedge\ldots\wedge w_p=w_p'$
for some $\theta\in \widehat{\gamma}$ and $\theta'\in \widehat{\gamma'}$.
Note in fact that the equalities $w_i=w_i'$ express the passing of values defined
via the evaluation $\rho'$ in the operational semantics.
\\
As a consequence,
$$
s(n_1,\ldots,n_k)~|~t(m_1,\ldots,m_u)
\rightarrow
s'(n_1',\ldots,n_k')~|~t'(m_1',\ldots,m_u')
$$
is a ground instance of one of the considered $MSR_{NC}$ rules.
\\
Thus, starting from the $MSR_{NC}$ configuration
$\widehat{G_l}({h_l})$, if we apply a rewriting step we obtain  a
new configuration in which $s(n_1,\ldots,n_k)$ has been replaced by $s'(n_1',\ldots,n_k')$,
and $t'(m_1',\ldots,m_k')$ has been replaced by $t(m_1',\ldots,m_u')$,
and all the other atomic formulas are as in $\widehat{G_l}({h_l})$.
This formula coincides with the definition of $\widehat{G_{l+1}}({h_{l+1}})$.

The proof of completeness is by induction on the length of an MSR run,
and by case-analysis on the application of the rules.
The structure of the case analysis is similar to the previous one
and it is omitted for brevity.
\end{proof}

\section{Verification of TDL Programs}
\label{Verification}\label{Safety}
Safety and invariant properties are probably the most important class of correctness
specifications for the validation of a concurrent system.
For instance, in Example \ref{AliceBob} we could be interested in proving
that every time a session terminates, two instances of thread $Init$ and $Resp$
have exchanged the two names generated during the session.
To prove the protocol correct independently from the number of names and threads
generated during an execution, we have to show that from the initial configuration
$G_0$ it is not possible to reach a configuration that violates the aforementioned
property.
The configurations that violate the property are those in which two
instances of $Init$ and $Resp$ conclude the execution of the protocol exchanging
only the first nonce. These configurations can be represented by looking at only two
threads and at the  relationship among their local data. Thus, we
can reduce  the verification problem of this safety property to
the following problem:
Given an initial configuration $G_0$ we would like to decide
if a global
configuration that {\em contains} {\em at least} two local
configurations having the form $\tuple{stop_A,i,n,m}$ and
$\tuple{stop_B,i',n',m'}$ with $n'=n$ and $m\neq m'$ for some
$i,i',n,n',m,m'$ is reachable.
This problem can be viewed as an extension of the {\em control
state reachability problem} defined  in \cite{AN00} in which we consider
both {control locations} and {local variables}.
Although control state reachability is undecidable
(see Corollary \ref{reachability}),
the encoding of TDL into MSR$_{NC}$ can be used to define a {sound} and {automatic}
verification methods for TDL programs.
For this purpose, we will exploit a verification method introduced for
\MSRC~in \cite{Del01,Del02}. In the rest of this section we will briefly
summarize how to adapt the main results in \cite{Del01,Del02} to the specific
case of MSR$_{NC}$.

Let us first reformulate the control state reachability problem
of Example \ref{AliceBob} for the aforementioned safety property
on the low level encoding into MSR$_{NC}$.
Given the MSR$_{NC}$ initial configuration $init$ we would like to
check that no configuration in $Post^*(\{init\})$ has the following
form
$$
\{stop_A(a_1,v_1,w_1),stop_B(a_2,v_2,w_2)\}\oplus \calM
$$
for $a_i,v_i,w_i\in \Rat~i:1,2$ and an arbitrary multiset of ground atoms $\calM$.
Let us call $U$ the set of {\em bad MSR$_{NC}$ configurations} having the
aforementioned shape.
Notice that $U$ is upward closed with respect to multiset inclusion, i.e.,
if $\calM\in U$ and $\calM\preccurlyeq\calM'$, then $\calM'\in U$.
Furthermore, for if $U$ is upward closed, so is $Pre(U)$.
On the basis of this property, we can try to apply the methodology proposed
in \cite{AN00} to develop a procedure to compute a finite representation $R$
of $Pre^*U)$.
For this purpose, we need the following ingredients:
\begin{enumerate}
\item a symbolic representation of upward closed sets of configurations
(e.g. a set of assertions $S$ whose denotation $\den{S}$ is $U$);
\item a computable symbolic predecessor operator $\bfPre$ working on sets of formulas
such that  $\den{\bfPre(S)}=Pre(\den{S})$;
\item a (decidable) entailment relation $Ent$ to compare
the denotations of symbolic representations, i.e., such that
$Ent(N,M)$ implies $\den{N}\subseteq \den{M}$.
If such a relation $Ent$ exists, then it can be naturally extended to sets
of formulas  as follows:  $Ent^S(S,S')$ if and only if for all $N\in S$ there exists $M\in S'$
such that $Ent(N,M)$ holds (clearly, if $Ent$ is an entailment, then
$Ent^S(S,S')$ implies $\den{S}\subseteq \den{S'}$).
\end{enumerate}
The combination of these three ingredients can be used to define
a verification methods based on backward reasoning as explained next.
\paragraph{Symbolic Backward Reachability}
Suppose that $M_1,\ldots,M_n$ are the formulas of our assertional language
representing the infinite set $U$ consisting of all bad configurations.
The {symbolic backward reachability procedure}  (SBR) procedure
computes a chain $\{I_i\}_{i\geq 0}$ of sets of assertions  such that
$$
\begin{array}{l}
I_0=\{M_1,\ldots,M_n\}\\
I_{i+1}=I_i~\cup~\bfPre(I_i)~~\hbox{for}~i\geq 0
\end{array}
$$
The procedure SBR stops when $\bfPre$ produces only redundant information, i.e.,
$Ent^S(I_{i+1},I_i)$.
Notice that $Ent^S(I_i,I_{i+1})$ always holds  since $I_i\subseteq I_{i+1}$.
\paragraph{Symbolic Representation}
In order to find an adequate represention of infinite sets of  MSR$_{NC}$
configurations  we can resort to the notion of {\em constrained configuration}
introduced in \cite{Del01} for the language scheme MSR($\calC$) defined
for a generic  constraint system $\calC$.
We can instantiate this notion with $NC$ constraints as follows.
A  {constrained configuration} over $\calP$ is a formula
$$
p_1(x_{11},\ldots,x_{1k_1})~|~\ldots~|~p_n(x_{n1},\ldots,x_{nk_n}):\varphi
$$
where $p_1,\ldots,p_n\in\calP$, $x_{i1},\ldots,x_{ik_i}\in\calV$
for any $i:1,\ldots n$ and $\varphi$ is an $NC$-constraint.
The denotation a constrained configuration
$M\doteq (\calM:\varphi)$  is defined  by taking the upward closure
with respect to multiset inclusion of  the set of ground instances, namely
$$
\den{M}=\{ \calM'~|~\sigma(\calM)\preccurlyeq \calM',~\sigma\in\Sol(\varphi)\}
$$
This definition can be extended to sets of MSR$_{NC}$ constrained
configurations with {\em disjoint variables} (we use variable renaming to avoid
variable name clashing) in the natural way.

In our example the following set $S_U$ of MSR$_{NC}$ constrained
configurations (with distinct variables) can be used to finitely
represent all possible violations $U$ to the considered safety property
$$
\begin{array}{ll}
S_U=\{ & stop_A(i_1,n_1,m_1)~|~stop_B(i_2,n_2,m_2)~:~n_1=n_2,m_1>m_2\\
      & stop_A(i_1,n_1,m_1)~|~stop_B(i_2,n_2,m_2)~:~n_1=n_2,m_2>m_1\}
\end{array}
$$
Notice that we need two formulas to represent $m_1\neq m_2$
using a disjunction of $>$ constraints.
The MSR$_{NC}$ configurations  $stop_B(1,2,6)~|~stop_A(4,2,5)$, and
$stop_B(1,2,6)~|~stop_A(4,2,5)~|~wait_A(2,7,3)$ are both
contained in the denotation of $S_U$.
Actually, we have that $\den{S_U}=U$.
This symbolic representation allows us to reason on infinite sets
of MSR$_{NC}$ configurations, and thus on global configurations of
a TDL program, forgetting the actual number  or threads of a given run.

To manipulate constrained configurations,  we can instantiate
to $NC$-constraints the {symbolic predecessor
operator} $\bfPre$ defined for a generic constraint system in \cite{Del02}.
Its definition is also given in Section \ref{PreOp} in Appendix.
From the general properties proved in \cite{Del02},
we have that when applied to a finite set of
MSR$_{NC}$ constrained configurations $S$, $\bfPre_{NC}$ returns a finite set
of constrained configuration  such that
$\den{\bfPre_{NC}(S)}=Pre(\den{S})$, i.e., $\bfPre_{NC}(S)$ is
a symbolic representation of the immediate predecessors of
the configurations in the denotation (an upward closed set) of $S$.
Similarly we can instantiate the generic entailment operator defined
in \cite{Del02} to MSR$_{NC}$ constrained configurations  so as to obtain
an a relation $Ent$ such that $Ent_{NC}(N,M)$ implies $\den{N}\subseteq \den{M}$.
Based on these properties, we have the following result.
\begin{proposition}
Let $\calT$ be a TDL program with initial global configuration $G_0$,
Furthermore, let $\calS$ be the corresponding MSR$_{NC}$ encoding.
and $S_U$ be the set of MSR$_{NC}$ constrained configurations denoting a given
set of bad TDL configurations.
Then, $init\not\in\bfPre_{NC}^*(S_U)$ if and only if
there is no finite run $G_0\ldots G_n$ and mappings $h_0,\ldots,h_n$ from the
names occurring in $G$ to non-negative rationals such that
$\TDLtoMSR{init}~\TDLtoMSR{G_0}({h_0})\ldots\TDLtoMSR{G_n}({h_n})$
is a run in $\calS$ and $\TDLtoMSR{G_n}({h_n})\in\den{U}.$
\end{proposition}
\begin{proof}
Suppose $init\not\in\bfPre_{NC}^*(U)$.
Since $\den{\bfPre_{NC}(S)}=pre(\den{S})$ for any $S$,
it follows that there cannot exist runs
$init \calM_0 \ldots \calM_n$ in $\calS$
such that  $\calM_n\in \den{U}$.
The thesis then follows from the Theorem \ref{soundcomp}.
\end{proof}
As discussed in \cite{BD02}, we have implemented our verification
procedure based on $MSR$ and {\em linear constraints} using a
CLP system with linear arithmetics.
By the translation presented in this paper, we can now reduce the
verification of safety properties of multithreaded programs
to a fixpoint computation built on {\em constraint operations}.
As example, we have applied our CLP-prototype to automatically verify
the specification of Fig. \ref{ABMSR}.
The unsafe states are those described in Section \ref{Verification}.
Symbolic backward reachability terminates after 18 iterations
and returns a symbolic representation of the fixpoint with 2590 constrained
configurations.
The initial state $init$ is not part of the resulting set.
This proves our original thread definitions correct with respect to
the considered safety property.
\subsection{An Interesting Class of TDL Programs}
\label{Decidable}
The proof of Theorem \ref{twocounters} shows
that verification of safety properties is undecidable for TDL
specifications in which threads have several local variables (they
are used to create linked lists). As mentioned in the introduction,
we can apply the sufficient conditions for the termination of the procedure SBR
given in \cite{BD02,Del02} to identify the following interesting subclass
of TDL programs.
\begin{definition}\rm
A monadic TDL thread definition $P=\tuple{Q,s,V,R}$ is such that $V$
is at most a singleton,  and every message template in $R$ has at most one variable.
\end{definition}
A monadic thread definition can be encoded into the monadic fragment
of MSR$_{NC}$ studied in \cite{Del02}. Monadic MSR$_{NC}$ specifications are
defined over atomic formulas of the form $p$ or $p(x)$ with $p$ is a predicate symbol
and $x$ is a variable, and on atomic constraints of the form $x=y$, and $x>y$.
To encode a monadic TDL thread definitions into a Monadic MSR$_{NC}$ specification,
we first need the following observation.
Since in our encoding we only use the constant $0$, we
first notice that we can restrict our attention to MSR$_{NC}$ specifications in which
constraints  have no constants at all.
Specifically,  to encode the generation of fresh names we only have to add an
auxiliary atomic formula $zero(z)$, and refer to it every time we need to express the
constant $0$. As an example, we could write rules like
$$
\arrowup{init}{}{fresh(x)~|~init_M(y)~|~zero(z)}~:~x>z,y=z
$$
for
initialization, and
$$
\begin{array}{l}
\arrowup{create(x)~|~zero(z)}{}
         {init_M(x')~|~init_{A}(id',n',m')~|~zero(z)}~:~\\
~~~~~~~~~~~~~~~~~~~~~~~~~~~~~~~~~~~~~~~~~~~~x'=x,id'=x,n'=z,m'=z,z'=z
\end{array}
$$
for all assignments involving the constant $0$.
By using this trick an by following the encoding of Section \ref{MSRTranslation},
the translation of a collection of monadic thread definitions directly returns
a {\em monadic} MSR$_{NC}$ specification.
By exploiting this property, we obtain the following result.
\begin{theorem}\label{termination1}
The verification of safety properties whose violations can be
represented via an upward closed set $U$ of global configurations is
decidable for a collection $\calT$ of monadic TDL definitions.
\end{theorem}
\begin{proof}
Let $\calS$ be the MSR$_{NC}$ encoding of  $\calT$
and  $S_U$ be the set of constrained configuration such that $S_U=U$.
The proof is based on the following properties.
First of all, the  MSR$_{NC}$ specification $\calS$ is monadic.
Furthermore, as shown in \cite{Del02}, the class of monadic MSR$_{NC}$ constrained
configurations is closed under application of the operator $\bfPre_{NC}$.
Finally, as shown in \cite{Del02}, there exists an entailment relation $CEnt$
for monadic  constrained configurations that ensures the
termination of the SBR procedure applied to a monadic MSR$_{NC}$ specification.
Thus, for the monadic MSR$_{NC}$ specification $\calS$,
the chain defined as $I_0=S_U$, $I_{i+1}=I_i\cup \bfPre(I_i)$
always reaches a point $k\geq 1$ in which $CEnt^S(I_{k+1},I_k)$,
i.e. $\den{I_k}$ is a fixpoint for $Pre$.
Finally, we note that we can always check for membership of
$init$ in the resulting set $I_k$.
\end{proof}
As shown in \cite{Sch02}, the complexity of verification methods based on
symbolic backward reachability relying on the general results in \cite{AN00,FS01}
is non primitive recursive.
%
%
%
\section{Conclusions and Related Work}
\label{conclusions}
In this paper we have defined the theoretical grounds for the possible application of
constraint-based symbolic model checking for the automated analysis of abstract models
of multithreaded concurrent systems providing name generation, name mobility, and unbounded control.
Our verification approach is based on an encoding into a low level
formalism based on the combination of multiset rewriting and constraints that allows
us to naturally implement name generation, value passing, and
dynamic creation of threads.
Our verification method makes use of symbolic
representations of infinite set of system states and of symbolic
backward reachability. For this reason, it can be viewed as a conservative
extension of traditional finite-state model checking methods.
The use of symbolic state analysis is  strictly related to the analysis methods based on
abstract interpretation. A deeper study of the connections with abstract interpretation
is an interesting direction for future research.

\paragraph{Related Work}
The high level syntax we used to present the abstract models of
multithreaded programs is an extension of the communicating finite state
machines used in protocol verification \cite{Boc78},
and used for representing  abstraction of multithreaded software
programs \cite{BCR01}. In our setting we enrich the formalism
with local variables,  name generation and mobility, and unbounded control.
Our verification approach is inspired by the recent work of Abdulla and Jonsson.
In \cite{AJ03}, Abdulla and Jonsson proposed an assertional
language for Timed Networks in which they use dedicated data
structures to symbolically represent configurations parametric in the
number of tokens and in the {\em age} (a real number) associated
to tokens. In \cite{AN00}, Abdulla and Nyl\'{e}n formulate a
symbolic algorithm using {\em existential zones} to represent
the state-space of Timed Petri Nets. Our approach generalizes the
ideas of \cite{AJ03,AN00} to systems specified via multiset
rewriting and with more general classes of constraints.
In \cite{AJ01}, the authors apply
similar ideas to (unbounded) channel systems in which messages can
vary over an infinite {\em name} domain and can be stored in a
finite (and fixed a priori) number of data variables.
However, they do not relate these results to multithreaded programs.
Multiset rewriting over first order atomic formulas has been proposed
for specifying security protocols by Cervesato et al. in \cite{CDLMS99}.
The relationships between this framework and concurrent languages based on
process algebra have been recently studied in \cite{BCLM03}.
Apart from approaches based on Petri Net-like models (as in \cite{GS92,BCR01}),
networks of {\em finite-state} processes can also be verified by means of
automata theoretic techniques as in \cite{BJNT00}.
In this setting the set of possible {\em local states} of individual
processes are abstracted into a {\em finite alphabet}. Sets of
global states are represented then as {\em regular languages}, and
transitions as relations on languages. Differently from the
automata theoretic approach, in our setting we handle
parameterized systems in which individual components have local
variables that range over {\em unbounded} values.
The use of constraints for the verification of concurrent systems is
related to previous works connecting Constraint
Logic Programming and verification, see e.g. \cite{DP99}. In this
setting  transition systems are encoded via CLP programs used to
encode the {\em global} state of a system and its updates. In the approach
proposed in \cite{Del01,BD02}, we refine this idea by using multiset rewriting
and constraints to {\em locally} specify updates to the {\em global} state.
In \cite{Del01}, we defined the general framework of multiset rewriting with
constraints and the corresponding symbolic analysis technique.
The language proposed in \cite{Del01} is given for a generic constraint system
$\calC$ (taking inspiration from $CLP$ the language is called
$MSR(\calC)$). In \cite{BD02}, we applied this formalism to verify properties
of mutual exclusion protocols (variations of the {\em ticket algorithm})
for systems with an arbitrary number of processes.
In the same paper we also formulated sufficient conditions for the termination
of the backward analysis.
The present paper is the first attempt of relating the low level language proposed
in \cite{Del01} to a high level language with explicit management of names and
threads.
\paragraph{Acknowledgments}
The author would like to thank Ahmed Bouajjani, Andrew Gordon, Fabio Martinelli,
Catuscia Palamidessi, Luca Paolini, and Sriram Rajamani and the anonymous reviewers
for several fruitful comments and suggestions.
\bibliographystyle{acmtrans}

\begin{thebibliography}{}

\bibitem[\protect\citeauthoryear{Abdulla, {C}er{\=a}ns, Jonsson, and
  Tsay}{Abdulla et~al\mbox{.}}{1996}]{ACJT96}
{\sc Abdulla, P.~A.}, {\sc {C}er{\=a}ns, K.}, {\sc Jonsson, B.}, {\sc and} {\sc
  Tsay, Y.-K.} 1996.
\newblock {General Decidability Theorems for Infinite-State Systems}.
\newblock In {\em {Proceedings 11th Annual International Symposium on Logic in
  Computer Science (LICS'96)}}. {IEEE Computer Society Press}, New Brunswick,
  New Jersey, 313--321.

\bibitem[\protect\citeauthoryear{Abdulla and Jonsson}{Abdulla and
  Jonsson}{2001}]{AJ01}
{\sc Abdulla, P.~A.} {\sc and} {\sc Jonsson, B.} 2001.
\newblock {Ensuring Completeness of Symbolic Verification Methods for
  Infinite-State Systems}.
\newblock {\em {Theoretical Computer Science}\/}~{\em 256,\/}~1-2, 145--167.

\bibitem[\protect\citeauthoryear{Abdulla and Jonsson}{Abdulla and
  Jonsson}{2003}]{AJ03}
{\sc Abdulla, P.~A.} {\sc and} {\sc Jonsson, B.} 2003.
\newblock {Model checking of systems with many identical timed processes}.
\newblock {\em {Theoretical Computer Science}\/}~{\em 290,\/}~1, 241--264.

\bibitem[\protect\citeauthoryear{Abdulla and Nyl{\'{e}}n}{Abdulla and
  Nyl{\'{e}}n}{2000}]{AN00}
{\sc Abdulla, P.~A.} {\sc and} {\sc Nyl{\'{e}}n, A.} 2000.
\newblock {Better is Better than Well: On Efficient Verification of
  Infinite-State Systems}.
\newblock In {\em {Proceedings 15th Annual International Symposium on Logic in
  Computer Science (LICS'00)}}. {IEEE Computer Society Press}, {Santa Barbara,
  California}, 132--140.

\bibitem[\protect\citeauthoryear{Ball, Chaki, and Rajamani}{Ball
  et~al\mbox{.}}{2001}]{BCR01}
{\sc Ball, T.}, {\sc Chaki, S.}, {\sc and} {\sc Rajamani, S.~K.} 2001.
\newblock {Parameterized Verification of Multithreaded Software Libraries}.
\newblock In {\em {7th International Conference on Tools and Algorithms for
  Construction and Analysis of Systems (TACAS 2001), Genova, Italy, April
  2-6,}}. {LNCS}, vol. 2031. {Springer-Verlag}, 158--173.

\bibitem[\protect\citeauthoryear{Bistarelli, Cervesato, Lenzini, and
  Martinelli}{Bistarelli et~al\mbox{.}}{2005}]{BCLM03}
{\sc Bistarelli, S.}, {\sc Cervesato, I.}, {\sc Lenzini, G.}, {\sc and} {\sc
  Martinelli, F.} 2005.
\newblock Relating multiset rewriting and process algebras for security
  protocol analysis.
\newblock {\em Journal of Computer Security\/}~{\em 13,\/}~1, 3--47.

\bibitem[\protect\citeauthoryear{Bochmann}{Bochmann}{1978}]{Boc78}
{\sc Bochmann, G.~V.} 1978.
\newblock Finite state descriptions of communicating protocols.
\newblock {\em Computer Networks\/}~{\em 2}, 46--57.

\bibitem[\protect\citeauthoryear{Bouajjani, Jonsson, Nilsson, and
  Touili}{Bouajjani et~al\mbox{.}}{2000}]{BJNT00}
{\sc Bouajjani, A.}, {\sc Jonsson, B.}, {\sc Nilsson, M.}, {\sc and} {\sc
  Touili, T.} 2000.
\newblock {Regular Model Checking}.
\newblock In {\em {Proceedings 12th International Conference on Computer Aided
  Verification (CAV'00)}}, {E.~A. Emerson} {and} {A.~P. Sistla}, Eds. {LNCS},
  vol. 1855. {Springer-Verlag}, {Chicago, Illinois}, 403--418.

\bibitem[\protect\citeauthoryear{Bozzano and Delzanno}{Bozzano and
  Delzanno}{2002}]{BD02}
{\sc Bozzano, M.} {\sc and} {\sc Delzanno, G.} 2002.
\newblock Algorithmic verification of invalidation-based protocols.
\newblock In {\em 14th International Conference on Computer Aided Verification,
  CAV '02}. Lecture Notes in Computer Science, vol. 2404. Springer.

\bibitem[\protect\citeauthoryear{Cervesato, Durgin, Lincoln, Mitchell, and
  Scedrov}{Cervesato et~al\mbox{.}}{1999}]{CDLMS99}
{\sc Cervesato, I.}, {\sc Durgin, N.}, {\sc Lincoln, P.}, {\sc Mitchell, J.},
  {\sc and} {\sc Scedrov, A.} 1999.
\newblock {A Meta-notation for Protocol Analysis}.
\newblock In {\em {12th Computer Security Foundations Workshop (CSFW'99)}}.
  {IEEE Computer Society Press}, {Mordano, Italy}, 55--69.

\bibitem[\protect\citeauthoryear{Delzanno}{Delzanno}{2001}]{Del01}
{\sc Delzanno, G.} 2001.
\newblock {An Assertional Language for Systems Parametric in Several
  Dimensions}.
\newblock In {\em {Verification of Parameterized Systems - VEPAS 2001}}.
  {ENTCS}, vol.~50.

\bibitem[\protect\citeauthoryear{Delzanno}{Delzanno}{2005}]{Del02}
{\sc Delzanno, G.} 2005.
\newblock {Constraint Multiset Rewriting}.
\newblock Tech. Rep. TR-05-08, Dipartimento Informatica e Scienze
  dell'Informazione, Universit\`a di Genova, Italia.

\bibitem[\protect\citeauthoryear{Delzanno and Podelski}{Delzanno and
  Podelski}{1999}]{DP99}
{\sc Delzanno, G.} {\sc and} {\sc Podelski, A.} 1999.
\newblock {Model checking in CLP}.
\newblock In {\em {Proceedings 5th International Conference on Tools and
  Algorithms for Construction and Analysis of Systems (TACAS'99)}}. Lecture
  Notes in Computer Science, vol. 1579. {Springer-Verlag}, Amsterdam, The
  Netherlands, 223--239.

\bibitem[\protect\citeauthoryear{Finkel and Schnoebelen}{Finkel and
  Schnoebelen}{2001}]{FS01}
{\sc Finkel, A.} {\sc and} {\sc Schnoebelen, P.} 2001.
\newblock {Well-Structured Transition Systems Everywhere!}
\newblock {\em {Theoretical Computer Science}\/}~{\em 256,\/}~1-2, 63--92.

\bibitem[\protect\citeauthoryear{German and Sistla}{German and
  Sistla}{1992}]{GS92}
{\sc German, S.~M.} {\sc and} {\sc Sistla, A.~P.} 1992.
\newblock {Reasoning about Systems with Many Processes}.
\newblock {\em {Journal of the ACM}\/}~{\em 39,\/}~3, 675--735.

\bibitem[\protect\citeauthoryear{Gordon}{Gordon}{2001}]{Gor00}
{\sc Gordon, A.~D.} 2001.
\newblock Notes on nominal calculi for security and mobility.
\newblock In {\em Foundations of Security Analysis and Design, Tutorial
  Lectures}. Lecture Notes in Computer Science, vol. 2171. Springer, 262--330.

\bibitem[\protect\citeauthoryear{Kesten, Maler, Marcus, Pnueli, and
  Shahar}{Kesten et~al\mbox{.}}{2001}]{Pnueli}
{\sc Kesten, Y.}, {\sc Maler, O.}, {\sc Marcus, M.}, {\sc Pnueli, A.}, {\sc
  and} {\sc Shahar, E.} 2001.
\newblock Symbolic model checking with rich assertional languages.
\newblock {\em Theoretical Computer Science\/}~{\em 256,\/}~1, 93--112.

\bibitem[\protect\citeauthoryear{Schnoebelen}{Schnoebelen}{2002}]{Sch02}
{\sc Schnoebelen, P.} 2002.
\newblock {Verifying Lossy Channel Systems has Nonprimitive Recursive
  Complexity}.
\newblock {\em {Information Processing Letters}\/}~{\em 83,\/}~5, 251--261.

\end{thebibliography}

\appendix
\section{Symbolic Predecessor Operator}
\label{PreOp}
Given a set of MSR$_{NC}$ configurations $S$, consider the
MSR$_{NC}$ {\em predecessor} operator
$Pre(S)=\{\calM|\calM\Rightarrow_{MSR}\calM',\calM'\in S\}$.
In our assertional language, we can define a symbolic version
$\bfPre_{NC}$ of $Pre$ defined on a set $\bfS$ containing
MSR$_{NC}$ constrained multisets (with {\em disjoint} variables)
as follows:
$$
\begin{array}{ll}
\bfPre_{NC}(\bfS)=\{~(\calA\oplus\calN~:~\xi)~~|~~
& (\calA\longrightarrow\calB : \psi)\in\calR,~~(\calM:\varphi)\in\bfS,\\
& \calM'\preccurlyeq\calM,~~\calB'\preccurlyeq\calB,\\
& (\calM':\varphi)~=_{\theta}~(\calB':\psi),~~\calN=\calM\ominus\calM',\\
& \xi\equiv (\exists x_1.\ldots x_k.\theta)\\
& \hbox{and}~x_1,\ldots, x_k~\hbox{are all variables not in}~\calA\oplus\calN\}.
\end{array}
$$
where $=_{\theta}$ is a matching relation between constrained configurations
that also takes in consideration the constraint satisfaction, namely
$$
(A_1~|~\ldots~|~A_n:\varphi)~=_{\theta}~(B_1~|~\ldots~|~B_m:\psi)
$$
provided  $m=n$ and there exists a permutation
$j_1,\ldots,j_n$ of $1,\ldots,n$ such that
the constraint
$\theta~=~\varphi\wedge\psi\wedge\bigwedge_{i=1}^n~A_i=B_{j_i}$
is {\em satisfiable}; here $p(x_1,\ldots,x_r)=q(y_1,\ldots,y_s)$
is an abbreviation for the constraints $x_1=y_1\wedge \ldots\wedge
x_r=y_s$ if $p=q$ and $s=r$, $false$ otherwise.

As proved in \cite{Del02},
 the symbolic operator $\bfPre_{NC}$ returns a set of MSR$_{NC}$
constrained configurations and it is correct and complete with
respect to $Pre$, i.e., $\den{\bfPre_{NC}(\bfS)}=Pre(\den{\bfS})$ for
any $\bfS$.
It is important to note the difference between $\bfPre_{NC}$ and a
simple backward rewriting step.

For instance, given the
constrained configurations $M$ defined as $p(x,z)~|~f(y):~z>y$ and the rule
$s(u,m)~|~r(t,v)\rightarrow p(u',m')~|~r(t',v')~:~u=t,m'=v,v'=v,u'=u,t'=t$
(that simulates a rendez-vous ($u,t$ are channels) and  value passing ($m'=v$)),
the application of $\bfPre$ returns
$s(u,m)~|~r(t,v)~|~f(y):~u=t,v>y$
as well as $s(u,m)~|~r(t,v)~|~p(x,z)~|~f(y):~u=t,x>y$
(the common multiset here is $\epsilon$).
\end{document}